\pdfoutput=1
\documentclass[10pt]{article}
\usepackage[preprint]{tmlr}

\usepackage{hyperref}
\usepackage{url}
\usepackage{graphicx}
\usepackage{rotating}
\usepackage{multirow}
\usepackage{amsmath,amssymb,amsfonts}
\usepackage{booktabs}
\usepackage{placeins}

\let\oldsubsection\subsection
\renewcommand{\subsection}{\FloatBarrier\oldsubsection}

\newcommand{\probeA}{Probe~A}
\newcommand{\probeB}{Probe~B}
\newcommand{\dA}{\Delta_{\mathrm{A}}}
\newcommand{\dB}{\Delta_{\mathrm{B}}}

\title{Saying, Not Knowing: 
Aggressively GGUF-Quantized Small Language Models Still Write Rare Words They Can No Longer Define}

\newcommand{\paperabstract}{

Post-training quantization to the GGUF format's mixed-precision K-quants is 
commonly how open-weight language models reach consumer hardware, yet its effect
on fine-grained lexical competence is uncharacterized. 
We audit
27 quantized artifacts across 13 families and four
architecture backbones, 0.35B--14B parameters, 
evaluated down their published ladder to Q2\_K (${\sim}2.6$ bits per weight), 
on 429 frequency-validated rare
English words under two probes: surface inclusion of a prompt-supplied word and
its one-sentence definition, scored by a tiered multi-synonym matcher,
its error measured by a blind LLM-judge census of every definition, with human verification.
Three regimes emerge at Q2: total collapse into unusable builds, 
severe semantic dissociation in sub-2B models, 
and mostly robust preservation above ${\sim}3$B.
In every sub-2B artifact, definitions fall $20$--$67\%$ below the artifact's baseline, 
typically several times the inclusion loss.
Two controls separate rarity from task difficulty:
within the rare set, loss rises with rarity in six of seven sub-2B artifacts, 
and on a 100-word common-word set
rare words lose more than common words in all eight, significantly in six.
Tokenizer vocabulary size does not predict the damage (Spearman
$\rho{=}0.12$); parameter count dominates ($\rho{=}0.72$), confirmed within five
of six same-tokenizer families.
Q4\_K\_M remains lexically clean at ${\geq}1$B.
The damage is frequency-graded, provider-dependent, and not calibrated by
WikiText-2 perplexity: across nine artifact-matched ladders, near-identical Q2
penalties ($44.7\%$/$47.6\%$) separate an artifact keeping its definitions
($3.6\%$) from one losing them ($43.6\%$).
Aggressively quantized small models can keep \emph{generating fluent text} while no longer \emph{knowing} what it means,
risking hardware-constrained deployments in domains where semantics carries consequences. Validation must be per artifact. 

}

\newpage

\author{\name Saurabh Kumar Singh \email saurabh.mam25@diat.ac.in \\
      \addr Department of Applied Mathematics, Defence Institute of Advanced Technology, Pune, India
      \AND
      \name Yogeshwar Singh Dadwhal \email yogeshwar\_singh@diat.ac.in \\
      \addr Department of Applied Mathematics, Defence Institute of Advanced Technology, Pune, India
      \AND
      \name Malhar Vedak \email malhar.pam26@diat.ac.in \\
      \addr Department of Applied Mathematics, Defence Institute of Advanced Technology, Pune, India}

\def\month{MM}
\def\year{YYYY}
\def\openreview{\url{https://openreview.net/forum?id=XXXX}}

\begin{document}

\maketitle

\begin{abstract}
\paperabstract
\end{abstract}

\section{Introduction}\label{sec:intro}

In self-hosted use, publicly available open-weight large language models (LLMs) 
are routinely deployed on consumer hardware after post-training quantization (PTQ), 
often as a default deployment format~\citep{ollamatags2026}.
GGUF is a binary file format from the \texttt{llama.cpp} ecosystem~\citep{llamacpp}
for storing quantized model weights and metadata, with K-quants as its family
of mixed-precision quantization schemes.
Served through tools such as Ollama,
it compresses weights from 16-bit precision (FP16, BF16) to as little as ${\sim}2.6$
bits per weight (Q2\_K), which allows the end user to
run 1B--14B models within 4--16\,GB of VRAM using hardware such as smaller workstation
and gaming GPUs.
This is not a niche pathway: we found that the Hugging Face Hub 
currently hosts over 200{,}000 GGUF builds~\citep{hfgguf2026}, and a 
single model repository of Llama-3.2 
reports over 81M pulls on Ollama's model library~\citep{ollamalibrary2026}. 
Millions of deployments, therefore, consist not of the full-precision model 
that the vendor released alongside its benchmarks, 
but rather of a third-party quantized artifact of it.
The default ``pull'' command used by Ollama, \texttt{ollama pull model\_name},
fetches a \texttt{q4\_K\_M} quantized variant of the Llama-3.2-3B, Mistral-7B, and Qwen2.5-7B models~\citep{ollamatags2026}, 
which are amongst the 27 artifacts used in this study. 
Model cards~\citep{mitchell2019modelcards} are the standard route by which vendors provide
performance and evaluation metrics about a model.
We are not aware of any standard model card
that reports those benchmarks for quantized artifacts alongside the
full-precision artifacts, or that reports them separately for lexical or rare-word
categories.
In this study we fill that gap in model evaluation, and present a failure mode
that is invisible to the aggregate benchmarks vendors release, and that the
WikiText-2 perplexity scores an AI practitioner deploying those models currently
trusts can register without identifying.

The motivation for this study has its origin in how humans use vocabulary in their
day-to-day life.
Human vocabulary follows a highly uneven frequency distribution.
A small set of very common words
accounts for much of the ordinary speech and writing, while the 
language itself contains a long tail of words that are individually rare.
This regularity in the frequency distribution of words is Zipf's
law~\citep{zipf1949}, and we use its frequency scale to 
select and stratify the rare-word set in Sec.~\ref{sec:words}.
Where a person encounters those rare words is not evenly distributed either.
Hayes and Ahrens counted how often words falling outside the ten thousand most common
types in English appeared across a broad range of spoken and written sources.
Adult books, popular magazines and newspapers carried roughly 53 to 68 such words for
every thousand running words of text.
Prime-time television carried about 23, and recorded conversation between college
graduates about 17~\citep{hayesahrens1988}.
Rare vocabulary is therefore substantially denser in print than in speech.
How much a person reads is in turn associated with the strength of their vocabulary and
general language ability~\citep{molbus2011}.
Differences in exposure also show up in what a person produces, specifically as the use
of the less frequent part of the vocabulary (rare words).
Laufer and Nation measured the share of a writer's output that is drawn from frequency
bands beyond the most common words.
That share distinguishes language learners at different levels of proficiency, 
and remains relatively stable when the same learner writes on a different topic~\citep{laufernation1995}.
The everyday form of this observation is that a well-read person is 
more likely to use a rare word like 
\emph{petrichor} instead of saying \emph{the smell after the rain}.  

If lexical knowledge in an LLM is encoded in a similar manner to humans, with
redundancy proportional to training exposure, then compressing the network
\emph{after} learning (through quantization) should erode the thinly-encoded long tail of the 
frequency distribution of the lexicon first. 
To test this initially, we devised a study to measure the vocabulary of the models,
by prompting models to define every word in a selected dictionary, and quantifying their 
failure to produce the required dictionary definitions, expecting
a more compressed model to hold ``less vocabulary'' compared to its full-precision counterpart.

Two corrections to that formulation shaped the current study, and both became
essential to the framework.

First, \emph{lexical knowledge is not the size of a word list}.
Research on human vocabulary has long treated knowing a word as having two dimensions
rather than one.
Anderson and Freebody named the first of these \emph{breadth}: ``the number of words
for which the person knows at least some of the significant aspects of meaning''.
The second they named \emph{depth}: ``the quality [\ldots] of understanding'' of any
one of those words~\citep{andersonfreebody1981}.
A count of the words someone knows is a measure of breadth. It says nothing about depth.

The same distinction is useful for language models, although for a different reason.
A definition is not a stored string that the model retrieves from a lookup
table~\citep{elhage2022superposition}.
An LLM composes distributed semantic associations into a new representation
at inference time in order to produce the required definitions. 
Knowing words, in the sense that matters, is the ability to
combine their meanings, so a single ``define it'' prompt confounds two separable
abilities: 1. producing a word form, and 2. reconstructing its meaning. 
To overcome this confound, we designed a two-probe study.

\begin{itemize}
\item \textbf{\probeA{}} asks the model to write sentences that naturally
include a rare word supplied in the prompt. Success is verbatim inclusion of the supplied word,
and we call the metric \emph{surface inclusion} because the word is given in the prompt.
\item \textbf{\probeB{}} asks for a one-sentence definition of the same word, scored against a
validated multi-synonym key.
\end{itemize}
The contrast between the two probes under identical compression levels,
across the breadth of parameter size and multiple model architectures and families,
over the same 429 words, is the paper's central object of study.

Both these probes are constructed from established practices for testing vocabulary.
Constructed-answer vocabulary tests have long asked subjects
to ``give a definition, a synonym, an illustration, or use the word in a sentence or
phrase''~\citep{andersonfreebody1981}.
In that tradition the two formats are alternatives to one another. 
A subject who succeeds at any one of them is credited with knowing the word, 
and the formats are pooled into a single score.
Our design keeps two of those formats apart and does not pool them.
Every word is put to every model in both formats, as the described \probeA{} and \probeB{},
at every level of compression, 
and what we report is the gap between the two results rather than their pooled sum.

Second, \emph{a model's vocabulary is not its tokenizer's word list}. 
How well a model commands a rare word should reflect how richly that word was encoded during the
pre-training process, by 
1. the model's exposure to it and 
2. the capacity available to store it with redundancy~\citep{allenzhu2025capacity}, not whether the 
word survives as a single token during tokenization.
The distinction is important because there is a competing account of where a model's
command over rare words lives, and that account makes a prediction we can directly test.
Frequency-driven subword tokenization tends to leave rare words split across 
more tokens than common words of comparable length~\citep{sennrich2016,kudo2018}.
Vocabulary size has separately been argued to be an under-allocated dimension of model
scaling, with compute-optimal vocabularies considerably larger than the ones most
current models use~\citep{tao2024vocab}.

Put together, these two observations suggest what we will call the \emph{fragmentation hypothesis}. 
If a model's command of a rare word depends on whether its tokenizer keeps
that word whole, then a model with a larger vocabulary should lose less rare-word
knowledge under quantization,
because fewer of its rare words were fragmented to begin with.
This hypothesis is directly testable from the data we collect and published vocabulary sizes of the models.

Our data show no support for this hypothesis: tokenizer vocabulary size
does not predict the model's
definitional degradation (Spearman $\rho{=}0.12$ across non-collapsed artifacts)
while parameter count is the dominant predictor of definitional robustness ($\rho{=}0.72$). 
This null result is not an isolated one.
Ali et al.\ trained 24 models at a 2.6B parameter scale, one for each of 24 tokenizers,
and tested whether the standard intrinsic measures of tokenizer quality predicted how
the resulting models performed on downstream tasks.
Most notable among those measures is \emph{fertility}, the average number of tokens a tokenizer
needs to represent a word.
They found such measures not consistently predictive, and described them as a questionable
proxy for downstream performance~\citep{ali2024tokenizer}.
Thus, tokenizer-level measures have already been shown to not necessarily
be able to predict model-level behaviour.

The intuition that a well-read reader commands rarer vocabulary is real in LLMs as well,
but it lives in exposure and capacity, not in the token inventory.
An almost-zero correlation on its own would only signify an absence of evidence, so we also
held the tokenizer fixed and varied the scale by changing parameter count.
Across the Qwen2.5 ladder, whose 0.5B, 1.5B and 7B artifacts share an identical 152k vocabulary,
definitional damage at Q2 still ranges from $-66.6\%$ at 1.5B to $-3.6\%$ at 7B
(Sec.~\ref{sec:vocab}). A larger tokenizer vocabulary did not protect a rare word
against definitional degradation in our sweep.

\begin{figure}[!t]
\centering
\includegraphics[width=0.75\textwidth]{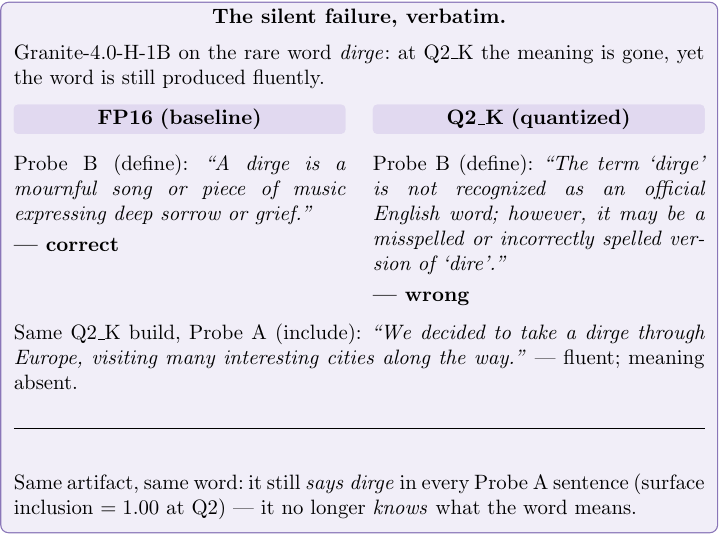}
\caption{Silent semantic failure, verbatim from the released outputs. Granite-4.0-H-1B
(Table~\ref{tab:main}) on the rare word \emph{dirge}, FP16 baseline vs.\ its Q2\_K build.
At FP16 the definition is correct; at Q2\_K the model reports the word as unrecognized,
yet still produces it fluently in every \probeA{} sentence (surface inclusion $=1.00$ at
Q2). In another run it is a medicine (``The doctor prescribed a dirge for her patient's
health issues''). Both census judges independently mark the Q2\_K definition wrong.
Quotes are verbatim; the word and full outputs are in the released
\texttt{definitions.csv} and \texttt{generations.csv} for Experiment~34}
\label{fig:silent}
\end{figure}

The stakes for the deployment of the model are dependent on the direction of the dissociation.
A user can notice when a model fails to produce rare words.
But a model that still produces a word fluently after losing its meaning fails
\emph{silently}: the output passes every eye test while the definitional knowledge 
it had at full precision is no longer there.
Specifically, we show that 1. this second failure mode is the one that actually occurs in small 
models under aggressive GGUF quantization, 2. it concentrates on 
low-frequency rare-word vocabulary, 3. the standard perplexity curve tracks it
but cannot calibrate it, and 4. it varies between quantized
builds of the same model. Figure~\ref{fig:silent} shows one such case verbatim.
\textbf{Fluency in producing a word is not evidence of meaning}.

Concretely, this paper contributes:
\begin{enumerate}

\item A two-probe, word-level paired approach, comparing production of a rare word 
in a sentence versus definitional knowledge of the same word under compression. 
The method employs a tiered, judge-validated scorer.  

\item An ecosystem audit of 27 quantized artifacts spread across 13 families, four
architecture backbones with added MoE variants for two architectures, varying 0.35B--14B in parameters
across public quantization ladders (216 file-level evaluations,
257 with the control runs of Sec.~\ref{sec:models}),
which yielded a three-regime taxonomy at Q2 as follows:
{\setlength{\leftmarginii}{2.2em}
\begin{enumerate}
\item artifact collapse into unusable builds;
\item selective semantic dissociation below ${\sim}2$B (with one
severe-but-symmetric 3.8B exception); and
\item robustness from ${\sim}3$B up;
\end{enumerate}}

\item A ladder-wide onset analysis showing definitional damage begins earlier in smaller
models, that Q4\_K\_M is lexically clean for every artifact of 1B or larger among the 27
while the aggressive Q4\_K\_S and Q3\_K\_S sub-variants already cost several
percent in some artifacts, and that at 3-bit
precision the damage concentrates in the S sub-variant, which accounts for 12
of the 22 significant 3-bit files;

\item Controls that distinguish word rarity from task difficulty: a 100-word 
common-word control across eleven artifacts, 
including every sub-2B model in the sweep, 
and a frequency dose-response gradient within the rare-word set;

\item Statistical treatment appropriate for the paired design: word-paired McNemar
tests, word-level Wilcoxon tests, per-model probe$\times$quantization
interactions with word-clustered robust standard errors,
Benjamini--Hochberg control over the full 436-comparison family, and
family-wise Holm--Bonferroni correction of the headline claims;

\item A test of the fragmentation hypothesis: tokenizer vocabulary
size does not predict definitional degradation (Spearman $\rho{=}0.12$) while
parameter count does ($\rho{=}0.72$), and a same-tokenizer scale ladder
(Qwen2.5 at 0.5B, 1.5B and 7B, all on an identical 152k vocabulary) separates
the two by holding the tokenizer fixed while damage varies with changing parameter size, with a
Mistral-lineage tokenizer contrast (Mistral-7B at 32k vs.\ Mistral-Nemo-12B at
131k, both robust) pointing the same way.

The cross-family correlation however confounds parameter count with model family. 
We therefore separate the two by replicating the comparison \emph{within} six
model families that each contribute multiple sizes on one tokenizer (18 of the
27 artifacts). 
The larger member is the less damaged one in five of the six families
(Sec.~\ref{sec:famscale}); and

\item Evidence from nine artifact-matched perplexity ladders (0.5B--7B,
including two robust controls, one unusable build, and a dual-build
pair) shows that WikiText-2 perplexity tracks the failure but cannot calibrate it. Perplexity
co-moves with overall damage severity, but supplies neither a threshold nor a
calibration: near-identical Q2 penalties of $+47.6\%$ and $+44.7\%$ separate an
artifact that loses $43.6\%$ of its definitions from one that loses just $3.6\%$.
The results also show that the effect depends on the specific builds being used, 
with Ollama and Unsloth artifacts of the same model producing different outcomes. 
This provides practical guidance for validating quantized artifacts as well. 
\end{enumerate}

\begin{figure}[!t]
\centering
\makebox[\textwidth][c]{\includegraphics[width=1.18\textwidth]{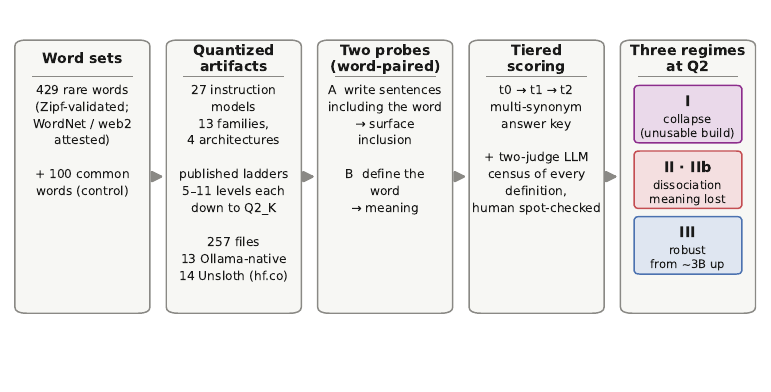}}
\caption{Study design and scale. Left to right: the stimulus sets (429 Zipf-validated
rare words; a 100-word common-word control); the 27 instruction-tuned quantized
artifacts (13 model families, four architecture backbones), each evaluated across its
published GGUF ladder (five to eleven levels per artifact), built by two providers (13
Ollama-native, 14 Unsloth via Hugging Face); with the build-source
and common-word controls, 257 file-level evaluations in total (221 rare-word, 36
common-word); the two word-paired
probes; the tiered scorer (t0/t1/t2), validated by a two-judge LLM census of every
definition and human spot-checked; and the three-regime outcome at Q2. 
In total the sweep produced $\approx$98{,}400 word-level
definition judgements and $\approx$295{,}200 inclusion generations ($\approx$393{,}600 scored outputs), all
on a single GPU within a 12--16\,GB VRAM budget (RTX~3060, 12\,GB; Kaggle T4, 16\,GB)}
\label{fig:pipeline}
\end{figure}

Figure~\ref{fig:pipeline} summarizes the study end to end.

\section{Related Work}\label{sec:related}

\paragraph{Quantization methods and their evaluation.}

Post-training quantization of large transformers became practical with
LLM.int8()~\citep{dettmers2022}. It holds a small number of outlier feature
dimensions in 16-bit and quantizes the rest of the multiplication to 8-bit. 
The weight-only methods that followed, GPTQ~\citep{frantar2023gptq} and
AWQ~\citep{lin2024awq}, leave activations untouched and compress only the stored weights. 
The GGUF K-quant family~\citep{llamacpp} uses
block-wise mixed-precision compression for consumer inference and deployment. It assigns
higher effective precision to more sensitive tensors, such as token-embeddings
and output matrices, and lower precision to other blocks, resulting in a reduction of the average
bits required to load the model in memory.
Systematic evaluations establish that 4-bit~\citep{whichquant2026}
weights generally can preserve aggregate performance with little degradation
on aggregate metrics while 2-bit~\citep{li2024evaluating} causes
substantially larger losses. The evaluations also show that the measured
result can depend on how model outputs are extracted and scored.
Our scorer uses multiple matching tiers to account for this.
That aggregate picture is not uniform across capabilities.
    \citet{zhou2026scaling} fitted a task-stratified scaling law to 293
quantized configurations of Qwen3 and Llama-3, with separate exponents for model
size, bit-width, group size and calibration-set size.
They found different bottlenecks for each capability: 
reasoning is limited by precision, application by model size, and 
memorization by calibration-set size. 
    \citet{wang2026compressed} studied a setting close to ours.
They tested factual recall in Llama3-8B and in the 7B and 14B members of
Qwen2.5, prompting each with subject-relation templates such as \emph{the
capital of [X] is}, and scoring whether the model returned the correct object.
They found two patterns.
Recall fell further in the smaller member of a family than in the larger one.
And quantization has little effect when the unquantized model is already reliable, 
but causes larger losses when its full-precision accuracy is lower.
Overall degradation was modest, and 4-bit models did not consistently perform worse than 8-bit models.
In relation to us, their index of fragility is the model's own full-precision accuracy, while ours is
corpus frequency, which was fixed before any model was run.
In the model-parameter range comparable to our study,
a pruning-versus-quantization comparison across six
models from 0.5B to 3.8B~\citep{zhou2025slm} recommends quantization as the safer
choice: at 4-bit it costs at most $5.6\%$ accuracy, while $75\%$ sparsity
pruning shows losses up to $48.07\%$.
Both studies went down to 4-bit quantization and used closed-form answers.
Our probes go below 4-bit in the ladder.

The limitation of aggregate-metric blindness is not unique to lexical evaluation. 
\citet{hooker2019forget,hooker2020bias} found that pruned and quantized vision
models showed comparable accuracy to full-precision counterparts, 
while diverging considerably on a
narrow, underrepresented subset of examples. 
Their affected subset and our rare words share a common characteristic: low frequency
in the training distribution.
Frequency does not always predict fragility in the same direction.
\citet{wu2026asymmetric} report the opposite direction for factual knowledge: in
Llama-3.1-8B, Qwen3-8B and Gemma-2-9B under eleven compression methods, questions about
popular entities lost more of their baseline accuracy than questions about rare ones.
Their frequency axis is entity popularity (Wikipedia page views) rather than word
frequency, and their models are 8--9B, a range where we also find robustness, so the two results do not conflict.
\citet{marchisio2024} showed Hooker's pattern across languages in a
multilingual LLM setting. In Japanese, automatic benchmarks recorded a $1.7\%$ average
drop, while human evaluators scoring realistic prompts reported $16.0\%$, an order of
magnitude larger.
The same gap has been measured for quantization alone, at scale, in social-bias
evaluation: 50 quantized models across 13 bias datasets~\citep{hua2026bias}.
Up to $21\%$ of individual responses flipped between biased and unbiased states
while the aggregate score did not change, and the flips were concentrated on the items
the full-precision model was least certain about.
Since the flips are bidirectional, the aggregate metrics remained 
unchanged even when individual responses changed. 
As an average cancels opposing changes, we compare each word with itself across
quantization levels instead of comparing group means (Sec.~\ref{sec:stats}).
For compressed LLMs specifically, the LLM-KICK benchmark~\citep{jaiswal2024kick}
shows that models compressed with negligible perplexity change can lose
significant \emph{knowledge-intensive} ability.
They used both pruning and quantization for compression, 
and their sharpest case for that blindness came from pruning.
We carry a narrower claim of perplexity changing with damage without 
indicating what exactly was damaged (Sec.~\ref{sec:ppl}).

\citet{duan2026} is closest to our method, where perplexity missed underlying feature damage
which they measured by degradation in active SAE features in weight-quantized models; and 
Hooker's is closest to our thesis: a behavioural, aggregate-metric-invisible subset effect. 
Our findings extend that pattern to lexical-semantic behaviour, showing that
the affected subset dissociates into separable surface and semantic
components.

These patterns also appear in applied deployment contexts. A systematic on-device
evaluation of GGUF quantization~\citep{ondevice2025} finds that
quantization tolerance depends sharply on scale: from Q5\_K to Q3\_K a 14B model holds
GSM8K accuracy at $89\%$ while a 1.5B model falls from $59.1\%$ to $46.5\%$, and code
generation is particularly sensitive to compression. Recent work documents disproportionate
loss of low-resource-language ability under aggressive quantization in machine
translation~\citep{mtforget2025}.
Our Regime~I collapses are not peculiar to our pipeline.
\citet{patodiya2026broken} audited 305 quantized artifacts in Ollama's official library
and found five that fail every functional test: four Qwen2.5-Coder-3B conversions and one
Phi-3.5-mini conversion. These are the same lineages as two of our three unusable builds
(Sec.~\ref{sec:collapse}), found independently with different tests. This supports our
reading that collapse is a property of individual published builds rather than of our probes.

\paragraph{Lexical knowledge probing.}
Prompt-based knowledge probing was introduced with LAMA~\citep{petroni2019}, which ranks
candidate fillers for a masked slot rather than generating text. Our definitional probe
(\probeB) keeps that same general probing approach but generates a full sentence, 
so we decode greedily to keep the output deterministic.
More recent work~\citep{ide2026} has formalized the approach of definitional generation and its
evaluation with lexicographic criteria and LLM-as-judge scoring.
We use LLM-as-judge scoring to validate our scorer rather than as the scorer, and
disclose the judges' identities~\citep{zheng2023judge}. 
The scorer itself is a tiered
matcher rather than a lexicographic criterion (Sec.~\ref{sec:scoring}).
Finally, subword tokenization~\citep{sennrich2016,kudo2018} is the motivation behind the
fragmentation hypothesis we test: rare words that are split into more subword tokens might
be more susceptible to definitional degradation as the model is quantized. Our data do not
support it (Sec.~\ref{sec:vocab}).

\paragraph{Compression and interpretable structure.}
\citet{duan2026} is the closest in spirit to our motivating question. 
Their study examines whether sparse-autoencoder features identified 
in a full-precision model survive round-to-nearest (RTN) quantization. 
It finds that feature degradation is graded,
and that a feature's full-precision statistics can predict its survival after quantization.
Importantly, perplexity can improve while a significant fraction 
of features actually degrade: on Gemma-2-2B, INT7 improves perplexity while 18.7\% 
of active features degrade. 
Their analysis is representational and uses RTN quantization.
In contrast, ours is behavioural and evaluates deployed GGUF artifacts. 
Sec.~\ref{sec:discussion} discusses how these 
findings relate to our results. 
To our knowledge, no prior work measures the surface-production/definition 
dissociation under quantization at the word level, across architecture classes, 
using the artifacts that users actually download.

\section{Methods}\label{sec:methods}

\subsection{Word sets}\label{sec:words}
The primary stimulus set contains 429 rare English words (nouns, verbs,
adjectives; e.g., \emph{pusillanimous}, \emph{gasconade}, \emph{recondite}).
Each word is paired with target synonyms.
Initially only one synonym was used, which was expanded later to multi-synonyms for our tiered scorer. 
One such word-synonym pair is \texttt{'allay':'soothe|ease|relieve|reduce|alleviate|lessen'}.
The candidate list was LLM-generated (Gemini 3.1 Pro) after corpus-scraped alternatives (Datamuse and WordNet-derived
lists) proved too noisy. The list was fixed before any model evaluation experiment was run.
The count 429 is the number of candidates the
generation step produced and which was retained unchanged. 
As we fixed the list before experiment runs, no words were added, dropped, or
substituted to favour a particular result. 
We have to prove the provenance of the LLM-generated rare-word set. We have to show that
the words are genuinely rare and genuinely real, independent of the model
that produced them. We establish these facts based on the two axes below.

To validate rarity independently, we used the \texttt{wordfreq}
corpus~\citep{speer2018}. The median Zipf frequency over the set is 2.34 (IQR 1.96--2.68, maximum
3.81), with no word reaching the Zipf${\geq}4$ band of everyday vocabulary.
Six of the 429 words returned a \texttt{wordfreq} Zipf frequency of 0.
Since \texttt{wordfreq} could assign 0 to both genuine rare words, and any fabricated
non-word, we had to show the 0 indicated the word falls below the corpus's detection floor rather
than that it is fabricated. To close this gap, each of the six
was checked independently against WordNet~\citep{miller1995} and the
public-domain \texttt{web2} dictionary word list.
Both the resources used for validation against fabrication are independent of \texttt{wordfreq} and the generating model,
and all six were confirmed to have a valid WordNet synset and a \texttt{web2} entry.
Applying all three checks (WordNet, \texttt{web2}, and a nonzero \texttt{wordfreq}
Zipf) to the entire list, all the 429 words in the set are attested by at least one
source independent of the generating model (425 in WordNet, 420 in \texttt{web2},
423 with nonzero Zipf). Every word in the set is independently verifiable and none is a model
fabrication.

The authenticity of the stimulus set therefore does not rest on
\texttt{wordfreq} or on the generating model. And for transparency we are releasing the full word list carrying
these per-word checks as columns so that readers can confirm this directly.
The rarity claim therefore rests on corpus statistics, independent of any LLM
judgment.

A 100-word common-word control was created as well,  
and its frequency checked against the same \texttt{wordfreq} corpus. 
The WordNet/\texttt{web2} fabrication checks were not run on this set,
since they exist to rule out fabrication risk specific to the LLM-generated rare list.
The set (median Zipf 4.97, minimum 3.74)
is nearly frequency-disjoint from the rare set (two rare words exceed the
lowest common-word frequency) and is used in Sec.~\ref{sec:control}. The disjoint nature
of the two sets allowed us to use the common-word set as a control against rare-word stimulus.

\subsection{Probes}\label{sec:probes}
\textbf{\probeA{} (surface inclusion).} The prompt is: \emph{``Write exactly
three grammatical sentences about \{topic\}. Naturally include the word
`\{word\}'. Do not define the word.''} The topic is drawn from a fixed topic
list by hashing (word, run) with MD5, so that topic assignment is
deterministic, resume-safe across interrupted runs, and independent of any
session state.
The ``do not define'' clause is to keep this probe disjoint from \probeB{}.
We ran each word three times, at temperature 0.3, to generate a total of
$429 \times 3 = 1{,}287$ outputs per build.
Success criterion is the case-insensitive word-boundary inclusion of the target
word in the output. 
We deliberately name this metric \emph{surface inclusion}, since the word is already
present in the prompt. So the probe is simply measuring whether output \emph{looks}
lexically intact, and not the word's competent (grammatical and semantic) usage 
in the generated sentence.
Its semantic quality is not scored (Sec.~\ref{sec:scoring}).

\textbf{\probeB{} (definitional knowledge).} The prompt is: \emph{``Define the
English word `\{word\}' in one concise sentence.''} This follows the general
knowledge-probe approach of \citet{petroni2019}, which used cloze-style prompts
to test a model's \emph{relational knowledge}. Because our probe generates a
sentence rather than ranking candidate fillers, we use greedy decoding
(by setting \texttt{temperature=0} in \texttt{Ollama}) to keep the output
deterministic.
\probeB{} thus produces one deterministic response per word ($n{=}429$),
which allowed us to do exact word-level
paired comparisons across quantization levels. 

\probeA{} is the easier, surface-level task by design, 
and we have deliberately kept this asymmetry in the study. 
The deployment-relevant failure we are trying to show is 
fluent surface inclusion of a word with degradation in semantics.
This study's claims rest on showing the \emph{contrast}
between the two probes under identical quantization levels, with the controls described below. The claims are not dependent on the two probes
having equivalent difficulty.

\subsection{Scoring and scorer validation}\label{sec:scoring}

\probeA{} scores verbatim (case-insensitive word-boundary) 
inclusion of the word supplied through the prompt. It does not 
judge if the word itself is used correctly in the model's output.
Judging correct usage requires expert
semantic and grammatical annotation rather than a deterministic scorer, which is
beyond the narrow scope of this study. 
We do not report it here and specify it as future work (Sec.~\ref{sec:future}).

\probeB{} responses are scored by a three-tier matcher in which each tier
was sequentially introduced to overcome non-intuitive results, produced by the tier below it.
Eventually, every response is scored under all three. 
Tier~0, which was the pilot instrument, only matched a single
canonical synonym as an exact word-boundary match. It was using the initial word-synonym 
pairing in the experiment files itself. 
It is blind to morphology (\emph{exhausted} counts, \emph{exhaustion} does not), and
it visibly undercounted correct definitions, producing full-precision baselines
of only 0.19--0.46. We flagged this scorer because some 
artifacts (Llama-3.2-3B, Gemma-2-9B)
moved in the  opposite direction on common-word control 
(definitional accuracy increased with increasing quantization), 
compared to the rare-word experiment.

Tier~1 keeps the single-synonym criterion but matches
under Snowball stemming~\citep{porter2001snowball}, so that \emph{exhausted},
\emph{exhaustion}, and \emph{exhausting} are scored the same.
This removed the morphological brittleness but still 
keeps rejecting any correct definition phrased
through a different synonym. This was flagged because of the difference between
\probeA{} inclusion success vs \probeB{} definitional success, where \probeA{} was above
0.90 for twenty of the twenty-four artifacts with a full-precision baseline, but \probeB{} started
below 0.57 for all twenty-four, and below 50 percent for nineteen of them.
We argued that models can't be that bad at definition, not at full precision at least.
So we checked the outputs and found that definitions were indeed being produced, but
single synonym was the culprit of low baseline score even at full precision. 

Tier~2 is the primary metric we finally used (all the results and conclusions are 
based on this scorer, not the previous tiers). 
It scores against a curated multi-synonym key (mean ${\sim}7$
acceptable synonyms or defining phrases per word)
under the same stemmed matching introduced in Tier~1, thus removing the paraphrase
brittleness as well. 
It raised the baselines at full precision to 0.62--0.94 (from 0.27--0.56 at Tier~1),
consistent with models genuinely knowing most of the words at full precision.
All three tiers are computed by the released scorer. Under the stricter Tier~0 the
headline sub-2B effects run in the same direction and are larger for seven of the
eight artifacts (all three tiers are in the released
word-level outcomes), so
Tier~2 is the final conservative choice, and its residual error is measured
directly by the LLM-judge census below. Our headline result holds under all three tiers.

Our validation logic treats the regex as the computationally cheap deterministic instrument and an
LLM judge as the expensive one, used to \emph{measure the cheap instrument's error}.
We use regex to keep the results and analysis deterministic, and validate its results using
an LLM, acting as \emph{LLM-as-a-judge}. 
Using an LLM as an evaluator of generated text is established
practice~\citep{zheng2023judge}, including specifically for judging
definition quality~\citep{ide2026}.

We ran a census in which LLM judges graded every definition in the study
(98{,}409 definitions from the 257 file-level \probeB{} evaluations of Sec.~\ref{sec:models}, each one artifact at one quantization level).
The primary judge is gpt-oss-20b (OpenAI, open-weight; \texttt{gpt-oss:20b}).
This judge was chosen for its independence: its developer built none of the
artifacts under study, and none of the models that generated the data under study.
A second judge, Gemma~4 (31B, Google; \texttt{gemma4:31b-cloud}), is used to replicate the census.
It is secondary because its developer (Google) also built three of the artifacts
under study, and the model (Gemini 3.1 Pro) that generated the word list.
Both judges also come from developers other than Anthropic, whose model
(Claude Fable~5) expanded the multi-synonym key (Sec.~\ref{sec:llmuse}),
so agreement between a judge and Tier~2 is not one model agreeing with itself.

Both judges were served through Ollama at temperature 0.
Each judge received only the word and the definition, so the artifact, the
quantization level and our scorer's verdict were hidden from it.
Definitions were sent in shuffled order, so that items judged together carry
no signal about artifact or quantization level to the judge.
Each judge graded every definition as either correct or incorrect under the same fixed rubric.

Our analysis finds that the Tier~2 regex is stricter than the primary judge.
It scores below gpt-oss in 247 of the 257 evaluations, and
10.1 points below on average, because a fixed synonym key can only credit the words it
lists, and so rejects a correct paraphrase that falls outside it.
The gap is systematic, and it largely cancels in a relative drop, on which our claims are based.
Across the 194 quantized evaluations that have a single,
non-zero full-precision baseline, the Tier~2 drop lies on average 2.5 points from the drop
measured by gpt-oss (median 1.6 points; 86.6\% of them within 5 points; $r = 0.984$).
If anything, Tier~2 slightly understates the definitional drop (mean signed error $+0.4$ points).

We then looked at where Tier~2 errs in the eight sub-2B artifacts, where our headline result lies. 
Treating a Tier~2 accept as a positive and the verdict of gpt-oss as the reference, 
a false negative is a correct definition that Tier~2 rejects, 
and a false positive is a wrong definition that Tier~2 accepts.
False negatives are the more common error at both baseline and Q2, but their share depends on the quantization level.
At baseline (all eight artifacts), 63.7\% of Tier~2 rejections are false negatives:
correct definitions phrased in words not present in our synonym key.
At Q2 (the seven with a Q2 build) only 25.6\% are, and the rest are genuine failures.
False positives run the other way: the precision of Tier~2 is 95.4\% at baseline but
78.7\% at Q2. 
So false negatives dominate at baseline, and false positives rise at Q2, where they raise the quantized score.
Under gpt-oss, four of the eight sub-2B drops are larger than under Tier~2 and four are
smaller, and none moves by more than 8.0 points, so the size of the headline drops does not depend on the scorer.
We report the Tier~2 results because they are deterministic.

The second judge, Gemma~4, agrees with gpt-oss on 96.2\% of definitions (Cohen's $\kappa = 0.81$) and confirms these results.
Tier~2 scores below Gemma~4 in 251 of the 257 evaluations, by 11.5 points on average,
and its drops lie on average 2.3 points from the drops measured by Gemma~4 ($r = 0.985$).
In the eight sub-2B artifacts, Gemma~4 finds the same error pattern as gpt-oss:
67.4\% of baseline rejections and 28.3\% of Q2 rejections are false negatives, and the
precision of Tier~2 is 97.1\% at baseline and 87.2\% at Q2.

We ran a blind human verification on a sample of 65 definitions,
stratified by regime (II and IIb, or III; Table~\ref{tab:main}),
quantization level (baseline, Q2) and Tier~2 verdict,
so it over-represents rejections (30 of 65, against 21.5\% in the full set).
Since we are dealing with rare words, the human reviewer first consulted three independent
dictionaries (\texttt{merriam-webster.com}, \texttt{dictionary.com} and
\texttt{dictionary.cambridge.org}),
and then judged whether each definition was correct, blind to the scorer's verdict.
All 65 appear in the census with identical output text.
The human agrees with gpt-oss on 62 of the 65, and with Gemma~4 on 62 as well
(Cohen's $\kappa{=}0.83$ for each).
The LLM judge therefore scores a definition much as the human reviewer does,
which is what lets it serve as the reference for Tier~2's error.

The human reviewer also checked the multi-synonym key.
A random 20-word sample of \texttt{multi\_synonyms.py} was verified against
standard definitions, independently of both the judges and the scorer, and
all 20 were verified as acceptable. 
Together, the two checks tie the pipeline to a human reading of the dictionary.

\subsection{Models and quantization ladder}\label{sec:models}

Table~\ref{tab:main} lists the 27 artifacts: 
13 model families, 
four architecture backbones 
(dense transformer; hybrid Mamba-2 Granite-4.0-H;
hybrid gated-convolution LFM2/LFM2.5; MatFormer Gemma~3n), 
two of which also appear in 
mixture-of-experts variants (LFM2-8B-A1B; Granite-4.0-H-7B),
spanning 0.35B--14B parameters and tokenizer vocabularies from 32k to 262k.

The models are grouped into 13 families, based on 
shared lineage, and tokenizer generation: 
Gemma-2, Gemma-3n, Llama-3.2, Mistral, Mistral-Nemo,
Ministral-3, Phi-3/3.5 (one lineage), 
Phi-4-mini (a distinct lineage with a
new 200k tokenizer), 
Qwen2.5, Qwen3, GLM-4, Granite-4.0-H, and LFM2/LFM2.5 (one lineage).

All artifacts were pulled and served through Ollama (using llama.cpp backend). 
They originate from two build providers: 13 are Ollama's native registry builds
and 14 are Unsloth-built GGUFs served through Ollama's Hugging Face
passthrough (\texttt{hf.co/unsloth} tags; marked $^{U}$ in the table).
Each artifact is evaluated at multiple quantization levels,
spanning the highest available (FP16/BF16 or Q8/Q6 when no
higher-precision file is published or the
full-precision file exceeds the VRAM budget, marked per row) down to
Q2\_K ($\ddagger$).
Across the 27 artifacts, we evaluate 216 builds
(artifact $\times$ its quantized versions we tested).
The exact ladder varies by artifact to reflect the sub-variants published by
each provider.

A ``level'' denotes a bit-class. When a level publishes multiple sub-variants
(K\_S/K\_M/K\_L), the ladders, the regime exemplars, and the perplexity runs
use the K\_M file (Q2 column: Q2\_K). The onset grid uses the worst
published sub-variant per level, and sub-variant-specific results are labelled
explicitly. Granite-4.0-H-350M ($\ddagger$) publishes no Q2 build and is analysed at Q3.

To probe artifact-source sensitivity directly, Llama-3.2-1B is evaluated in
both providers' builds: its Ollama build is one of the 27 primary artifacts, and its
Unsloth build, the build-source control, is used only to compare providers.
Reasoning-distilled models were deliberately excluded
from the study because chain-of-thought (CoT) token (e.g., 
\texttt{DeepSeek-R1} emits CoT tokens as \texttt{<think>} and \texttt{</think>})
leakage at small scale can affect both probes, hence
Qwen3-4B was evaluated in its non-thinking instruct variant (2507).
Experiments were run on a single RTX~3060 (12\,GB) and Kaggle T4 instances
(16\,GB), a 12--16\,GB VRAM budget throughout.

The released evaluation corpus contains 257 file-level \probeB{} evaluations
spanning every precision evaluated in the sweep: 221 rare-word evaluations
from the 27 artifacts' ladders (216 files) plus the 5 files of the
build-source control, and 36 common-word control evaluations
(eleven artifacts). These correspond to
$\approx$98{,}400 word-level definition outcomes. The paired \probeA{} runs
add $\approx$295{,}200 inclusion generations, for a total of $\approx$393{,}600
probe outputs across the study.

The full evaluation ran in $\approx$155 hours (${\sim}6.4$ days) of measured
single-GPU wall-clock time, within our 12--16\,GB VRAM budget. 

An exploratory pilot phase was used to select the models and configuration for full sweeps.
It was used to validate the pipeline and smoke test our setup.
Those pilot results are not included in the reported results. 
All reported numbers come from the subsequent neutral sweep with $N{=}429$.

\paragraph{Unusable builds.} 

We define a quantized artifact as an \emph{unusable build} in this study
when its accuracy on \emph{both} probes falls to near
zero against its own baseline. 
It neither produces the target word on \probeA{} 
nor defines it in \probeB{}. 
Because the two probes fail together, such an artifact
retains no usable capability on either task.
There is no rare-word-selective
interpretation of such a failure possible in our analysis.
The label describes the behaviour of one file at one precision, not
of the model it was built from, whose other builds may be unaffected.
We make no claim about why that file behaves as it does. Artifacts matching
this description form Regime~I of Table~\ref{tab:main}; \emph{unusable build},
\emph{collapsed build}, and \emph{Regime~I artifact} refer to the same classification.

\subsection{Statistical analysis}\label{sec:stats}

Each quantized level is compared against its own baseline. 
\probeB{} outcomes are paired at word level 
(same 429 words, deterministic decoding), 
so the primary test is McNemar's on discordant pairs~\citep{mcnemar1947}, 
using the continuity-corrected $\chi^2$ form and the exact binomial form where a
discordant cell is zero; the common-word comparisons use the exact form throughout.
Because the \probeA{}'s three runs per word are not independent,
we aggregate them to per-word success fractions and use the Wilcoxon signed-rank
test~\citep{dror2018}. 
The dissociation claim itself is tested per model with a
logistic probe$\times$quantization interaction using word-clustered robust
standard errors,
which are reported in Table~\ref{tab:main}.

Several statistics we report sit outside both multiple-testing correction
families described below: the rank correlations of Sec.~\ref{sec:vocab} and
Sec.~\ref{sec:ppl}, and the two selectivity controls of Sec.~\ref{sec:control},
which are per-artifact logistic models with word-clustered robust standard
errors, a word-type$\times$quantization term for the common-word contrast and a
continuous Zipf term for the frequency gradient.
The Benjamini--Hochberg family consists of word-paired per-file probe comparisons,
and the Holm--Bonferroni family of the nine pre-specified interactions.
Adding a differently-grained test to either would change interpretations of those corrections.
The interaction p-values we report for artifacts outside that nine-member
family are likewise unadjusted.
Corrected by Benjamini--Hochberg within each control, the six significant common-word
interactions and the six significant gradients all remain significant.

The dissociation claim is evaluated primarily through the two selectivity controls, not on the probe$\times$quantization interaction test
above: the common-word contrast and the frequency gradient operate entirely
\emph{within} \probeB{}, holding scale fixed and varying only word frequency,
so cross-probe scale differences cannot arise. The interaction test is
corroborating evidence, reported as odds ratios (Sec.~\ref{sec:regime2})
rather than p-values alone so effect magnitude can be judged directly. We
scope it as corroborating because, with \probeA{} near ceiling and \probeB{}
in mid-range, an ordinal (non-crossover) interaction significant on the
log-odds scale can in principle be attenuated under a different monotone
transformation of accuracy. This is the classic ``removable interaction''
critique of dissociation claims, and exactly the scale-dependence the
within-\probeB{} controls above are built to avoid.

Two layers of multiplicity control are applied. 
First, across the full family of
every evaluated non-baseline file at every level under both probes,
we control the false-discovery
rate (FDR) by Benjamini--Hochberg~\citep{bh1995} at $q{=}0.05$. 
The family contains a total of 436 paired comparisons:
1. Wilcoxon on per-word success fractions for
\probeA{}; and 
2. McNemar on word-paired Tier-2 outcomes for \probeB{}, under the
same convention as Table~\ref{tab:main}.
It covers both the rare-word ladders, and
the common-word control files. Of the total 127
raw-significant results, 95 remain significant after the correction.

Second, the headline dissociation claims are corrected family-wise by
Holm--Bonferroni~\citep{holm1979} over a prespecified family fixed by cohort definition: 
the probe$\times$quantization interaction of
every sub-2B artifact at its most aggressive level used in the study (Q2 for the
seven artifacts that publish a Q2 build, Q3 for Granite-4.0-H-350M, which
publishes none), together with the transitional Qwen2.5-3B at Q3, its last functional level, for nine interactions in total. 
Eight of the nine interactions remain significant after the correction.
The sole casualty is LFM2-700M ($p{=}.084$ against the final Holm threshold
of $.05$). It is the one family member that was already individually
non-significant and is treated as marginal throughout.
Exact p-values are reported throughout.

\subsection{Use of large language models}\label{sec:llmuse}

Following our reading of the AMEE Guide No. 192 on AI use in research~\citep{cleland2026disclosure},
we document our use of LLMs here.

LLMs appear inside the experimental pipeline in three roles, whose primary models come from
three different developers:

1. Gemini 3.1 Pro (Google) generated the rare-word candidate list, whose rarity was then
validated against corpus statistics (\texttt{wordfreq}) and its existence against WordNet and \texttt{web2};
2. Claude Fable~5 (Anthropic) expanded the multi-synonym answer key, which was human spot-checked; and
3. gpt-oss-20b (OpenAI) judged every definition to validate the scorer, with Gemma~4
(Google) as a replication judge and a human-verified sample as the anchor.

Human verification results for the judges and the synonym key are reported above. 
Outside the pipeline, Claude models (Anthropic; Fable~5 through Cowork, and Opus~5,
Sonnet~5 and Opus~5.5 through Claude Code) helped write the analysis code, including the
Tier~2 scorer and the paired tests, and assisted with editing and refinement of the text.
They also identified gaps in our evidence, which led us to extend the perplexity ladders and
the common-word control. ChatGPT (OpenAI), Gemini (Google) and Claude were used to discuss
the design and results early in the project. 
The authors reviewed, revised and verified all of it.
All scientific claims, the final text, and responsibility for the presented work
rest with the authors.

Proposing candidate material with a model, then gating inclusion on separate
criteria, follows established practice: \citet{wang2023selfinstruct} generate candidate items with a model and
filter the resulting pool for diversity and validity before use; \citet{heyman2023} find that ChatGPT produces norming-quality lexical
data comparable to human output on a related task; and \citet{gilardi2023} find LLM annotation matching or exceeding human
crowd-worker agreement. In our pipeline the generating model only proposes
the candidate words. Each retained word must independently clear the corpus-frequency and lexical-existence checks above before it
is used in the experiments.

\begin{sidewaystable}
\caption{Rare-word performance at the most aggressive published K\_M level (Q2\_K
unless noted) vs.\ each artifact's baseline. $\dA$, $\dB$: relative change in
\probeA{} (surface inclusion) and \probeB{} (definition, Tier-2). $p_B$:
McNemar on word-paired \probeB{}. Int.\ $p$: probe$\times$quantization
interaction, word-clustered; ``(+)'' marks positive coefficients (definition
better preserved than inclusion), ``floor'' means unidentifiable with both
probes near zero, ``n.e.'' not estimable due to separation (baseline inclusion
1.00). Granite-4.0-H-350M publishes no Q2 build; its row ($\ddagger$) reports
Q3\_K\_M. Deltas are computed from unrounded accuracies}
\label{tab:main}
\centering
\footnotesize
\setlength{\tabcolsep}{3.5pt}
\begin{tabular}{@{}llrrl rr rr l@{}}
\toprule
Model & Class & Params & Vocab & Base & $A_{\mathrm{base}}{\to}Q2$ & $\dA$(\%) & $B_{\mathrm{base}}{\to}Q2$ & $\dB$(\%) & $p_B$ / Int.~$p$\\
\midrule
\multicolumn{10}{@{}l}{\emph{Regime I --- artifact collapse (global, both probes)}}\\
Qwen2.5-3B & dense & 3.1B & 152k & FP16 & $.92{\to}.00$ & $-100.0$ & $.89{\to}.00$ & $-100.0$ & $10^{-115}$ / floor\\
Phi-3-3.8B & dense & 3.8B & 32k & FP16 & $.96{\to}.02$ & $-98.3$ & $.92{\to}.03$ & $-97.2$ & $10^{-116}$ / floor\\
Phi-3.5-3.8B & dense & 3.8B & 32k & FP16 & $.97{\to}.00$ & $-99.6$ & $.94{\to}.02$ & $-98.0$ & $10^{-119}$ / floor\\
\midrule
\multicolumn{10}{@{}l}{\emph{Regime II --- selective dissociation (inclusion preserved, meaning lost)}}\\
Granite-4.0-H-350M$^{U\ddagger}$ & hyb-SSM & 0.35B & 100k & FP16 & $.75{\to}.76$ & $+0.9$ & $.62{\to}.32$ & $-47.3$ & $5{\times}10^{-22}$ / $8{\times}10^{-16}$\\
Qwen2.5-0.5B & dense & 0.5B & 152k & FP16 & $.60{\to}.71^{\dagger}$ & $+17.0^{\dagger}$ & $.68{\to}.54$ & $-20.3$ & $6{\times}10^{-7}$ / $3{\times}10^{-13}$\\
LFM2-700M$^{U}$ & hyb-conv & 0.7B & 65k & FP16 & $.93{\to}.85$ & $-8.9$ & $.69{\to}.40$ & $-41.8$ & $10^{-20}$ / $.084$\\
Granite-4.0-H-1B$^{U}$ & hyb-SSM & 1.0B & 100k & FP16 & $.91{\to}.89$ & $-2.1$ & $.84{\to}.48$ & $-43.6$ & $10^{-28}$ / $6{\times}10^{-15}$\\
LFM2-1.2B$^{U}$ & hyb-conv & 1.2B & 65k & FP16 & $.95{\to}.92$ & $-2.4$ & $.72{\to}.51$ & $-29.8$ & $6{\times}10^{-15}$ / $.005$\\
LFM2.5-1.2B$^{U}$ & hyb-conv & 1.2B & 65k & FP16 & $.97{\to}.93$ & $-4.1$ & $.72{\to}.37$ & $-49.0$ & $10^{-27}$ / $.013$\\
\midrule
\multicolumn{10}{@{}l}{\emph{Regime IIb --- severe on both probes, definition worse}}\\
Llama-3.2-1B & dense & 1.2B & 128k & FP16 & $.89{\to}.58$ & $-34.3$ & $.82{\to}.32$ & $-61.1$ & $10^{-43}$ / $.004$\\
Qwen2.5-1.5B & dense & 1.5B & 152k & FP16 & $.78{\to}.53$ & $-32.5$ & $.87{\to}.29$ & $-66.6$ & $6{\times}10^{-51}$ / $5{\times}10^{-15}$\\
Phi-4-mini-3.8B$^{U}$ & dense & 3.8B & 200k & FP16 & $.94{\to}.59$ & $-36.7$ & $.91{\to}.52$ & $-43.3$ & $5{\times}10^{-33}$ / $.68$ (symm.)\\
\midrule
\multicolumn{10}{@{}l}{\emph{Regime III --- robust through Q2 (all $|\dB|\leq14.1\%$)}}\\
Gemma-2-2B & dense & 2.6B & 256k & FP16 & $.92{\to}.90$ & $-2.3$ & $.89{\to}.81$ & $-9.4$ & $5{\times}10^{-5}$ / $.0502$\\
Granite-4.0-H-3B$^{U}$ & hyb-SSM & 3.0B & 100k & FP16 & $.93{\to}.95$ & $+1.5$ & $.90{\to}.82$ & $-8.1$ & $.002$ / $8{\times}10^{-4}$\\
Ministral-3-3B$^{U}$ & dense & 3.0B & 131k & FP16 & $.95{\to}.92$ & $-3.4$ & $.92{\to}.90$ & $-2.3$ & n.s. / $.26$\,(+)\\
Llama-3.2-3B & dense & 3.2B & 128k & FP16 & $.96{\to}.92$ & $-4.5$ & $.89{\to}.77$ & $-14.1$ & $9{\times}10^{-8}$ / $.59$\\
Qwen3-4B$^{U}$ & dense & 4.0B & 152k & FP16 & $.98{\to}.98$ & $+0.6$ & $.90{\to}.81$ & $-9.8$ & $4{\times}10^{-5}$ / $.007$\\
Gemma-3n-E4B$^{U}$ & MatFormer & 4.0B & 262k & BF16 & $1.00{\to}.99$ & $-1.2$ & $.91{\to}.83$ & $-8.9$ & $10^{-5}$ / n.e.\\
Granite-4.0-H-7B$^{U}$ & hyb-MoE & 7.0B & 100k & Q8 & $.97{\to}.95$ & $-2.4$ & $.90{\to}.91$ & $+1.3$ & n.s. / $.006$\,(+)\\
Mistral-7B & dense & 7.2B & 32k & FP16 & $.91{\to}.90$ & $-1.2$ & $.82{\to}.78$ & $-5.7$ & $.018$ / $.33$\\
Qwen2.5-7B & dense & 7.6B & 152k & FP16 & $.95{\to}.91$ & $-4.1$ & $.90{\to}.87$ & $-3.6$ & n.s. / $.23$\,(+)\\
Ministral-3-8B$^{U}$ & dense & 8.0B & 131k & BF16 & $.98{\to}.96$ & $-2.2$ & $.92{\to}.93$ & $+1.5$ & n.s. / $.003$\,(+)\\
LFM2-8B-A1B$^{U}$ & MoE & 8.3B & 65k & FP16 & $.97{\to}.94$ & $-3.5$ & $.88{\to}.80$ & $-9.0$ & $8{\times}10^{-5}$ / $.46$\,(+)\\
Gemma-2-9B & dense & 9.2B & 256k & FP16 & $.99{\to}.98$ & $-0.9$ & $.89{\to}.87$ & $-1.8$ & n.s. / $.24$\,(+)\\
GLM-4-9B & dense & 9.4B & 151k & FP16 & $.96{\to}.96$ & $-0.7$ & $.87{\to}.81$ & $-6.5$ & $.008$ / $.36$\\
Mistral-Nemo-12B & dense & 12.2B & 131k & Q8 & $.94{\to}.92$ & $-3.0$ & $.86{\to}.79$ & $-8.2$ & $.002$ / $.83$\\
Ministral-3-14B$^{U}$ & dense & 14.0B & 131k & Q6 & $.98{\to}.98$ & $-0.4$ & $.96{\to}.94$ & $-2.4$ & n.s. / $.40$\\
\bottomrule
\multicolumn{10}{@{}l}{\footnotesize $^{\dagger}$Inflated by a repetition
effect: at Q2, this 0.5B artifact inserts the target word mechanically
(${\geq}3$ repetitions in 43\% of outputs vs.\ 25\% at FP16).}\\
\multicolumn{10}{@{}l}{\footnotesize $^{U}$Unsloth-built GGUF served through
Ollama's Hugging Face passthrough (\texttt{hf.co/unsloth}); unmarked artifacts are Ollama-native registry builds.}\\
\multicolumn{10}{@{}l}{\footnotesize $^{\ddagger}$No Q2 build is published for
this artifact; its row reports Q3\_K\_M, the most aggressive K\_M level;
Q3\_K\_S is more aggressive.}\\
\end{tabular}
\end{sidewaystable}

\section{Results}\label{sec:results}

\subsection{Three regimes at Q2}\label{sec:regimes}
\begin{figure}[p]
\centering
\includegraphics[width=\textwidth]{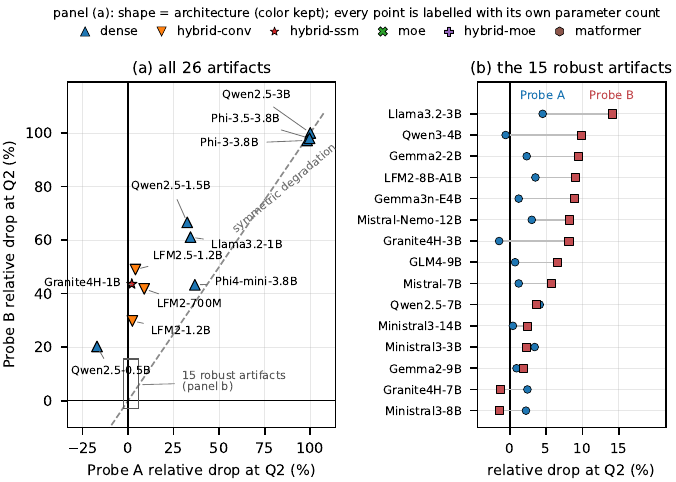}
\caption{Relative drop at Q2 on both probes, for the 26 primary artifacts with a
published Q2 build ($\ddagger$). (a) \probeA{} (surface inclusion, $x$)
against \probeB{} (definition, $y$). Regime-II artifacts hug the $y$-axis except the
three IIb members (Llama-3.2-1B, Qwen2.5-1.5B, Phi-4-mini-3.8B), whose larger
inclusion loss shifts them rightward; the collapse cohort sits on the diagonal
near 100\%; robust models cluster at the origin. The Qwen2.5-0.5B point at
negative $x$ is the repetition effect of Table~\ref{tab:main}. (b) The fifteen
artifacts inside that origin cluster, one row each}
\label{fig:scatter}
\end{figure}

Table~\ref{tab:main} and Fig.~\ref{fig:scatter} summarize the core result:
at the most aggressive precision, artifacts do not follow a single pattern of degradation.
They separate into three regimes: 
1. total collapse on both probes (Sec.~\ref{sec:collapse}), 
2. selective dissociation in which inclusion is
preserved while definitions degrade (Sec.~\ref{sec:regime2}), and
3. relative robustness on both probes.

Table~\ref{tab:main} additionally separates a subgroup of the
dissociation regime, labelled IIb: 
artifacts whose surface inclusion also
falls substantially while definitions degrade even more strongly. 
Two of its members
(Llama-3.2-1B, Qwen2.5-1.5B) remain significantly definition-selective under
the interaction test and are therefore classified with Regime~II;
the third member (Phi-4-mini, 3.8B) is included because of the severity of its losses
even though its damage is symmetric.
IIb is the severe end of the dissociation
regime, rather than a separate fourth regime.

\begin{figure}[p]
\centering
\includegraphics[width=\textwidth]{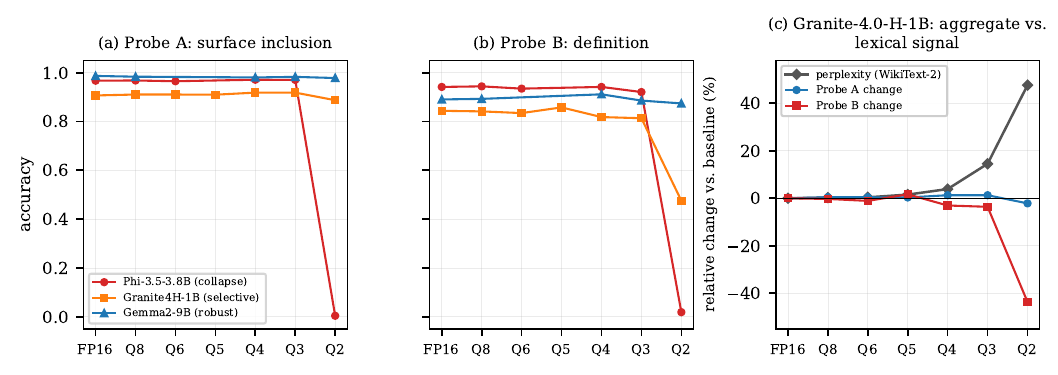}
\caption{(a,b) Accuracy across the quantization ladder for one exemplar of each
regime (K\_M files). Surface inclusion is flat through Q3 for all three; only
the definitional probe separates the regimes. (c) For the selective exemplar
(Granite-4.0-H-1B), WikiText-2 perplexity (diamonds) degrades smoothly while the
definitional probe (squares) declines discontinuously at Q2 and inclusion (circles)
barely moves}
\label{fig:signatures}
\end{figure}

Figure~\ref{fig:signatures} shows one exemplar ladder for each regime.
Surface inclusion is unchanged through Q3 for all three examples (K\_M files).
The regimes are separated only by the
definitional probe.
And the collapse is like a cliff at Q2, instead of a gradual slope.

\citet{zhou2026collapse} give a mechanistic account of this kind of
failure, separating it into two modes:
\emph{signal degradation},
where the computational pattern remains intact and
quantization error accumulates as noise; and 
\emph{computation collapse}, where components stop functioning and the knowledge
signal is destroyed in the early layers. In that mode, even patching clean
full-precision activations into the build failed to restore the correct answer.
Their collapsed 2-bit builds produced chaotic content such as meaningless numbers
and repetitive loops. 
Our Regime~I shows the same behaviour, including the non-linguistic
digit runs we recover from the Qwen2.5-3B Q2 build,
while Regimes~II and~III are consistent with their signal degradation mode.
They also reported that the two modes are not strictly tied to a particular
bit-width but reflect the nature of the damage, which matches our
finding that regime membership is a property of the individual artifact rather
than of its quantization level.

\subsection{Regime I: collapse is an artifact property}\label{sec:collapse}

Three artifacts meet the unusable-build criterion of Sec.~\ref{sec:models} at
Q2: Phi-3, Phi-3.5, and Qwen2.5-3B. 
Each loses over $97\%$ on \emph{both} probes
against their own FP16 baselines (Table~\ref{tab:main}).
Two further observations support the classification on individual members: 
1. Manual inspection of all 1{,}287 Qwen2.5-3B Q2 generations finds only non-linguistic
output (digit runs, mixed-script fragments; mean length 412${\to}$133
characters), and 
2. for Phi-3.5, the common-word control collapses alongside the
rare-word probes ($0.95{\to}0.03$ inclusion, $0.87{\to}0.03$ definition;
Sec.~\ref{sec:control}), indicating that its failure is not limited to 
just the rare-word vocabulary. 
The classification is based only on the paired probe results; the above
two observations only provided the corroborating evidence.

A third, instrument-based check provided the final confirmatory evidence for the classification:
WikiText-2 perplexity for the Qwen2.5-3B Q2 build measures 21{,}797 against
a 9.04 full-precision baseline (${\sim}2{,}400{\times}$;
Sec.~\ref{sec:ppl}), against 11--46 for every functional artifact in
our perplexity sweep at Q2. Thus, the build is unusable by measurement using a staple instrument (perplexity score), 
not just by our two probes.

All three are official registry builds that users can pull without any warning on the
download page.
We therefore report these separately from the selective semantic effects, where the artifacts can 
retain useful output while losing definitional performance.

Our experiments provide no explanation for the cause of this collapse.
We cannot say whether the quantization of
these specific files went wrong, or whether these models are intrinsically not
compressible to ${\sim}2.6$ bits per weight.
The two possibilities are indistinguishable by their probe outputs alone.

The three models also reach the Q2 failure differently.
For the Phi pair (Q3\_K\_M variant) the drop is abrupt: definitional loss is ${\leq}4\%$ at Q3. 
Qwen2.5-3B instead fails through an intermediary stage. 
Its Q3 outputs are still fluent English (manual inspection;
non-ASCII rate 0.03\%) with inclusion only moderately reduced
($0.92{\to}0.83$ at Q3\_K\_M) while definitions fall $47\%$
($0.89{\to}0.47$; McNemar $p{=}4{\times}10^{-34}$; interaction
$p{=}5{\times}10^{-8}$, odds ratio $0.27$).

Notably, for Qwen2.5-3B, this stage is also visible in the chosen staple instrument: 
its WikiText-2 perplexity rises
$+70.1\%$ at Q3, the largest Q3 penalty in our sweep.
So, perplexity does register
that this artifact is damaged, but not what exact capability is actually 
increasing the perplexity score. The following section sheds more light on this shortcoming: 
perplexity alone cannot reveal that the damage is definitional.

This is a genuine selective dissociation at 3.1B,
the largest we observed outside the sub-2B cohort, before becoming unusable at Q2.
This provides
another example of the continuum described below and
suggests that the Qwen2.5 lineage is particularly fragile under this format.

\subsection{Regime II: selective dissociation below ${\sim}2$B}\label{sec:regime2}

Every sub-2B artifact ($\ddagger$) shows a larger loss in definitional accuracy at Q2
compared to its corresponding surface inclusion. 
This holds across dense (Llama-3.2-1B, Qwen2.5-0.5B/1.5B),
hybrid-SSM (Granite-4.0-H-1B), and hybrid-convolution (LFM2/LFM2.5)
architectures. 

Two of these, Llama-3.2-1B and Qwen2.5-1.5B, are placed in stratum IIb of
Table~\ref{tab:main} because their surface inclusion also drops by roughly a third.
Their probe$\times$quantization interactions nonetheless remain strongly definition-selective. 
We therefore treat IIb as the severe end of Regime~II rather than a separate regime.

Phi-4-mini (3.8B) degrades severely but symmetrically ($p{=}.68$):
severe, not selective.

The clearest example is Granite-4.0-H-1B: surface inclusion barely
moves ($-2.1\%$, word-level Wilcoxon $p{=}.046$) while definitional accuracy
falls $43.6\%$ (McNemar $p{=}1.2{\times}10^{-28}$; 179 vs.\ 21 discordant words).
The probe$\times$quantization interaction is significant in six of
the seven sub-2B artifacts whose Q2 build is used in the study (Table~\ref{tab:main}).
LFM2-700M is the exception among these seven, with a marginal interaction
($p{=}.084$) because its \probeA{} also declines.
The eighth sub-2B model, Granite-4.0-H-350M, has no Q2 build.
Its Q3 endpoint appears in Table~\ref{tab:main} ($\ddagger$) and is discussed below.

The probe$\times$quantization interaction (Sec.~\ref{sec:stats})
applies a logistic regression to predict probe success based on probe 
identity (A or B), quantization level (baseline or Q2), and their interaction. 
The interaction coefficient evaluates whether quantization alters 
the \emph{odds of success} for \probeB{} by a distinctly different multiplicative 
factor than for \probeA{}, controlling for any shared, proportional performance decline. 
This relationship is quantified by the \emph{Odds Ratio} ($e^{\text{coefficient}}$), 
which expresses the relative magnitude of quantization's multiplicative 
effect on \probeB{}'s odds compared to \probeA{}'s. 
An odds ratio of 1 indicates the absence of a differential effect, signifying
both \probeA{} and \probeB{} move proportionally, in either direction, or not move at all. 

For Granite-4.0-H-1B, the odds ratio is 0.21 (Wald 95\% CI: 0.14--0.31). 
This shows that the impact of quantization on the odds of a correct 
definition is approximately one-fifth of its impact on the odds of surface inclusion, 
demonstrating a selectively differential impairment rather than a uniform decline. 
We report \emph{Odds Ratio} in this study instead of raw percentage-point differences 
to systematically account for baseline performance disparities. 
Since \probeA{} is above 0.90 for twenty of the twenty-four artifacts that have
a full-precision baseline, its potential for absolute
percentage-point decline is constrained on an absolute percentage scale.
The log-odds scale inherently corrects for this ceiling effect 
to provide a more robust scale to measure the degradation.

The same pattern holds across the cohort: the six significant interactions 
correspond to odds ratios between 0.19 and 0.58.
At Q2, the multiplicative change in the odds of producing a
correct definition is between roughly one fifth and one half ($\sim\times3$) of the corresponding
change in surface-inclusion odds. 

Every individually significant interaction in this cohort survives the 
pre-specified nine-member family-wise Holm--Bonferroni correction. 
This includes the six sub-2B interactions at Q2, 
Granite-4.0-H-350M's interaction at Q3, 
and Qwen2.5-3B's interaction at Q3. 
The interaction for LFM2-700M is excluded from this surviving group, 
as it was only marginally significant prior to the correction.

The regimes are descriptive strata over what is plausibly a continuum, and
the full interaction column supports this interpretation. 
Qwen2.5-3B, which sits at the regime boundary, demonstrates this directly. 
It shows selective dissociation at Q3,
then collapses at Q2: the same artifact is moving
along the continuum as compression increases.
Selective decline remains statistically detectable in some models in the 3--4B range,
although the effect is milder compared to smaller models.
Granite-4.0-H-3B ($p{=}8{\times}10^{-4}$) and Qwen3-4B ($p{=}.007$) show significant 
interactions, while Gemma-2-2B is
marginal at $p{=}.0502$. 
Llama-3.2-3B provides a related, but distinct case.
Its rare-word-selective definitional decline ($\dB{=}{-}14.1\%$,
$p_B{=}9{\times}10^{-8}$) is statistically significant, but its probe interaction is
not individually significant at this model size ($p{=}.59$).

No artifact at 7B or above shows a significant definition-selective interaction.
The only significant 7B+ interactions are positive, indicating that definitions are better preserved than inclusion. 
The effect therefore concentrates in the smaller artifacts of our sweep.
The largest artifact with a significant definition-selective interaction is Qwen3-4B;
among the 7B--14B artifacts, no definition-selective interaction is detected.
It is not a universal law, and our roster cannot speak to sizes above 14B.

\subsection{Damage onset across the ladder}\label{sec:onset}

The smallest evaluated artifact provides the sharpest illustration of
\emph{scale-dependence} in quantization-induced definitional damage.
Granite-4.0-H-350M shows a first, modest definitional dip already at Q4
($-12.5\%$, McNemar $p{=}.001$), well before quantization becomes
aggressive. 
By Q3 that dip has become severe semantic dissociation: definitions
fall $47.3\%$ at Q3\_K\_M ($0.615{\to}0.324$, McNemar $p{=}5{\times}10^{-22}$)
and $60.2\%$ at the more aggressive Q3\_K\_S, while surface inclusion moves
only $+0.9\%$. 
The probe$\times$quantization interaction at Q3 is
significant ($-1.24$, $p{=}8{\times}10^{-16}$, odds ratio $0.29$), confirming
the measured gap between the two probes is not there by chance.

Its common-word control for the same artifact confirms the loss is, indeed, rare-selective. 
Common-word definitions are unchanged at Q3 ($-2.9\%$,
McNemar $p{=}.82$) while rare-word definitions fall $47.3\%$
(word-type$\times$quantization interaction $-1.11$,
$p{=}3.5{\times}10^{-6}$).

\begin{figure}[p]
\centering
\includegraphics[width=\textwidth]{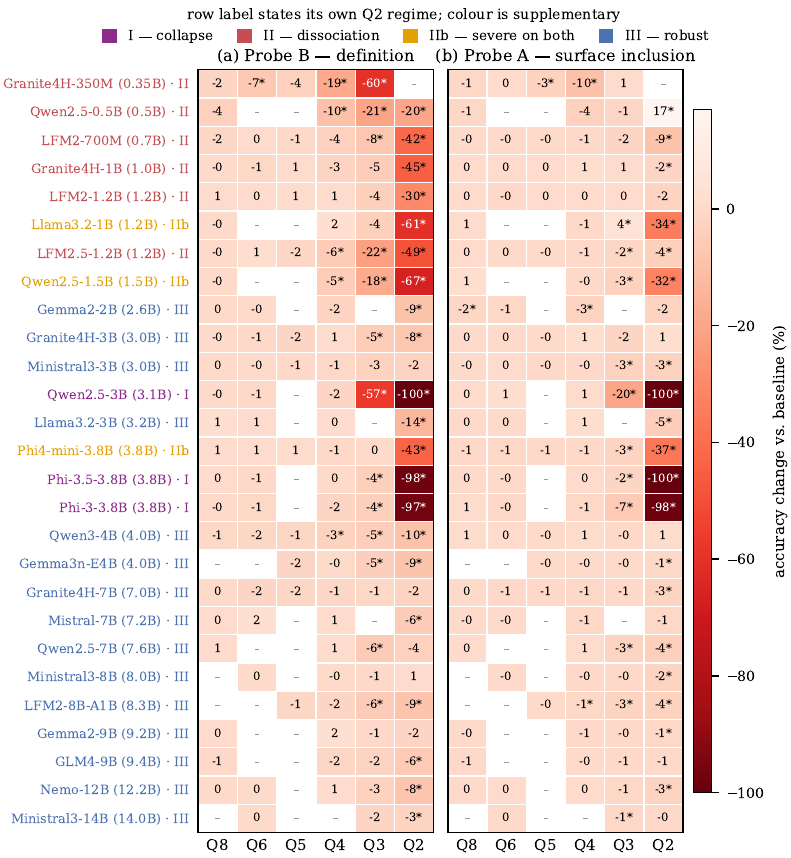}
\caption{Accuracy change against baseline at every evaluated precision level, both
probes, showing the worst sub-variant per level, models ordered by parameter
count (``--'' marks a level not evaluated). (a) \probeB{} (definitional,
Tier-2) for all 27 artifacts, 216 files; asterisks mark raw per-comparison
McNemar $p{<}.05$, and 30 of the 40 starred cells survive study-wide
Benjamini--Hochberg correction. (b) \probeA{} (surface inclusion); asterisks
mark word-level Wilcoxon $p{<}.05$. Severe inclusion losses occur only in the
three Regime-I builds and the three Regime-IIb artifacts; the $+17$ cell is
the Qwen2.5-0.5B repetition effect of Table~\ref{tab:main}}
\label{fig:onset}
\end{figure}

We present the full analysis in the grid in Fig.~\ref{fig:onset}, which shows
this selective dissociation is a general pattern in small models
at aggressive quantization levels. It first shows up in Granite-4.0-H-350M as it's
the smallest model (by number of parameters, not disk space) in our study.
The smaller the model, the earlier in the quantization ladder
definitional loss begins. While the entire sub-2B cohort shows severe degradation at Q2,
the degradation starts at higher precision levels.

The Q4 is the empirically \emph{safe} boundary against this degradation for larger models.  
The standard Q4\_K\_M file maintains lexical competence for every primary artifact of 1B or
larger size, while the aggressive Q4\_K\_S already causes small significant
losses (3--6\%) in three of the artifacts: LFM2.5-1.2B, Qwen2.5-1.5B, and Qwen3-4B.
Below 1B the picture diverges: the two smallest artifacts lose
8--19\% at either Q4 variant (Granite-4.0-H-350M $-12.5\%$/$-18.9\%$,
Qwen2.5-0.5B $-7.6\%$/$-10.0\%$, all four significant), while LFM2-700M
shows no significant loss at either.

At Q3, 22 of the 51 studied 3-bit files show significant loss.
While the loss is generally minor (4--8\%) in most models and concentrated in the Q3\_K\_S
sub-variant, it's already large in LFM2.5-1.2B ($-22\%$) and
Granite-4.0-H-350M ($-60\%$) at Q3\_K\_S level. 
(Note: The 51-file count includes control-ladder files; 
restricting to the 27 primary artifacts returns 50 files, 
21 of which show significant loss).

The selection of a sub-variant (K\_L, K\_M, K\_S) at a particular quantization level
matters as well.
At 3-bit precision the damage is concentrated in the
S sub-variant, which accounts for 12 of the 22 significant files.
Although the trend is dominant, it is not universal. 
Among the thirteen artifacts with a
significant 3-bit file and both sub-variants studied,
S is the more damaging file in eleven, while the two
exceptions, LFM2-700M and Phi-3.5-3.8B, degrade more at
K\_M and K\_L respectively.
Which sub-variant degrades more is therefore a
property of the individual builds being used, rather than of their bit-class.
The results are consistent with the artifact-source sensitivity of Sec.~\ref{sec:source}.

\begin{figure}[p]
\centering
\includegraphics[width=0.87\textwidth]{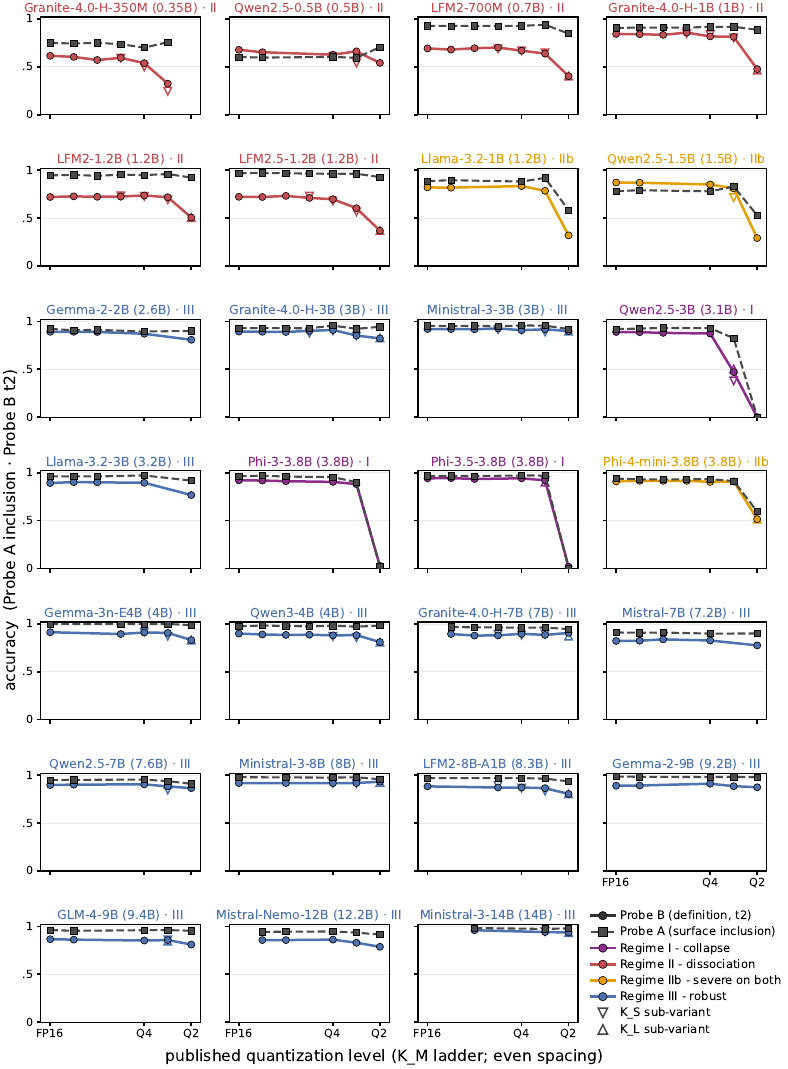}
\caption{Absolute probe trajectories across the quantization ladder for all 27
artifacts, ordered by parameter count. \probeB{} (definitional accuracy,
Tier-2) is a solid line with circle markers; \probeA{} (surface
inclusion) is a dashed line with square markers. On the \probeB{} line,
filled circles are the standard K\_M file at each level and open markers the
K\_S ($\triangledown$) and K\_L ($\triangle$) sub-variants where published.
Colour gives the Q2 regime of Table~\ref{tab:main}; Granite-4.0-H-350M
publishes no Q2 build and ends at Q3.
Qwen2.5-0.5B and Qwen2.5-1.5B are the only artifacts whose \probeA{} baseline is 
below their \probeB{} baseline, causing their lines to cross. 
The rise in \probeA{} for Qwen2.5-0.5B at Q2 reflects the repetition effect reported in Table~\ref{tab:main}
}
\label{fig:gallery}
\end{figure}

Fig.~\ref{fig:onset} summarizes the change from the baseline of an artifact, and
Fig.~\ref{fig:gallery} plots the absolute trajectory of both probes for every one of the
27 artifacts across its evaluated ladder.
The K\_M ladder remains essentially flat until the final precision rungs of the small models,
where it falls steeply. It collapses to the floor in the three Regime-I builds
and falls $43.3\%$ in Phi-4-mini-3.8B, the one Regime-IIb artifact at this size. 
Across the remaining artifacts it declines only
gradually from ${\sim}3$B up, the largest such loss being Llama-3.2-3B's
$-14.1\%$ at Q2.

\subsection{Rare-word selectivity: common-word control and frequency gradient}\label{sec:control}

It is intuitive to ask whether defining a word is simply harder than including
it in a sentence; so, any compression should damage our \probeB{} more. %
We proactively controlled for this variable of
\emph{task-difficulty} through two independent analyses.

\begin{figure}[!t]
\centering
\includegraphics[width=0.9\textwidth]{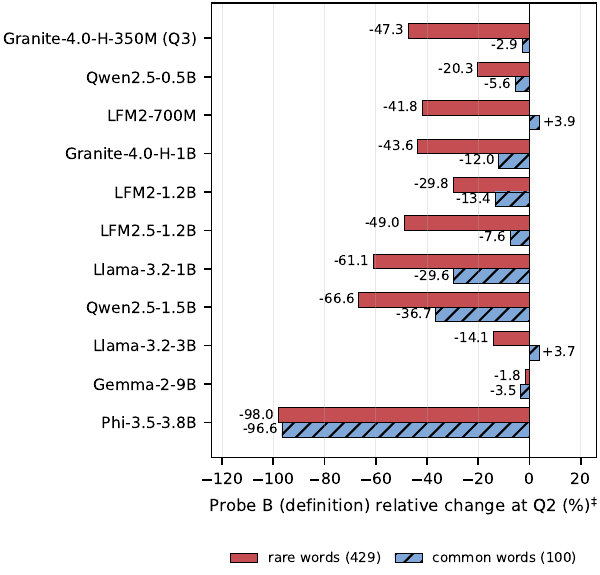}
\caption{Common-word control: relative change in definitional accuracy (\probeB,
Tier-2) at Q2 ($\ddagger$) for eleven artifacts, solid bars for the 429 rare
words and striped bars for the 100 common words. Word-type$\times$quantization
interactions are given in the text}
\label{fig:control}
\end{figure}

\textbf{Common-word control.} Eleven artifacts, comprising every sub-2B model in the
sweep plus the robust and collapse exemplars,
were re-evaluated on the 100 common words with an identical pipeline
(Fig.~\ref{fig:control}). 

\emph{Regime-II (rare-selective damage):} under Tier~2, the change in common-word
definitions at Q2 is not significant in three Regime-II artifacts, while rare-word
definitions decline sharply; the word-type$\times$quantization interaction tests the
difference directly.

Granite-4.0-H-1B: common $-12.0\%$ (McNemar $p{=}.076$) vs.\
rare $-43.6\%$ (word-type$\times$quantization interaction $-1.19$,
$p{=}4{\times}10^{-4}$).
LFM2.5-1.2B: common $-7.6\%$ ($p{=}.33$) vs.\
rare $-49.0\%$ (interaction $-1.17$, $p{=}2{\times}10^{-4}$).
LFM2-700M: common $+3.9\%$ ($p{=}.65$) vs.\ rare $-41.8\%$ (interaction $-1.38$,
$p{=}1.4{\times}10^{-6}$).
(Note: interactions are word-clustered, with negative values indicating rare-selective
damage. Granite-4.0-H-350M, which has no Q2 build, shows the same signature at Q3; 
it is the sixth significant interaction counted below.)

The two Regime-IIb artifacts in this cohort lose common-word
definitions as well:
Llama-3.2-1B $-29.6\%$ ($p{=}4{\times}10^{-5}$), and
Qwen2.5-1.5B $-36.7\%$ ($p{=}2{\times}10^{-6}$), which is 
consistent with their severe-on-both-probes classification.
Yet their rare-word loss remains significantly larger 
(interactions $-1.11$, $p{=}4{\times}10^{-4}$; $-1.48$, $p{=}5{\times}10^{-6}$).

In the remaining two sub-2B artifacts the
interaction is directionally consistent but not individually significant.
Qwen2.5-0.5B $-0.39$, $p{=}.12$, and LFM2-1.2B $-0.30$, $p{=}.32$. 
The latter does have a significant common-word dip of $-13.4\%$, $p{=}.043$. 
The broader pattern of word-type interaction is significant in six of the eight sub-2B artifacts,
and no sub-2B artifact shows the reverse pattern. 
For larger models and Regime~I artifacts: at Q2, Llama-3.2-3B's
rare-word definitions drop $14.1\%$ ($p{=}9{\times}10^{-8}$) while
common-word definitions are unchanged ($+3.7\%$, McNemar $p{=}.55$);
Gemma-2-9B shows no degradation on either word class at any level; and
Phi-3.5 collapses on both.

Task difficulty alone cannot explain this effect:
within the same task and the same artifact, damage is concentrated on
low-frequency vocabulary.

\begin{figure}[!t]
\centering
\includegraphics[width=0.65\textwidth]{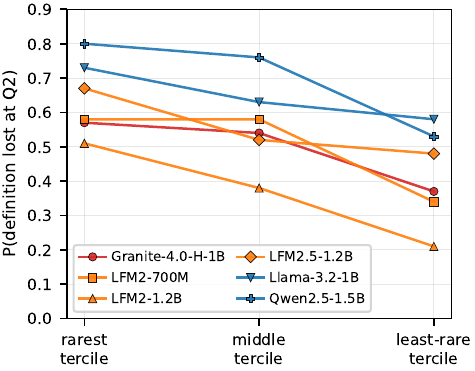}
\caption{Frequency dose-response within the rare-word set. 
Tracks words defined correctly at the baseline to show the probability of 
definitional loss at Q2, grouped by Zipf-frequency tercile.
Line colour = architecture class, marker = model. 
Loss probability rises with rarity in all six (logistic Zipf coefficients $-0.51$
to $-1.10$, all $p{\leq}.013$).
Qwen2.5-0.5B (not shown) is frequency-flat}
\label{fig:gradient}
\end{figure}

\textbf{Frequency gradient within the rare set.} 
The same conclusion is further validated from frequency selectivity inside the 429-word set itself,
using its Zipf range (0--3.81). 
Conditioning strictly on words defined correctly at baseline, the
probability that a definition is lost at Q2 rises significantly with rarity
in six of the seven sub-2B artifacts with a Q2 build (Fig.~\ref{fig:gradient}). 
Rarest vs.\ least-rare tercile loss rates: Granite-4.0-H-1B $0.57$ vs.\
$0.37$ (logistic Zipf coefficient $-0.59$, $p{=}.002$); LFM2-700M $0.58$
vs.\ $0.34$ ($-0.83$, $p{<}.001$); LFM2-1.2B $0.51$ vs.\ $0.21$
($-0.96$, $p{<}.001$); LFM2.5-1.2B $0.67$ vs.\ $0.48$ ($-0.60$,
$p{=}.003$); Qwen2.5-1.5B $0.80$ vs.\ $0.53$ ($-1.10$, $p{<}.001$);
Llama-3.2-1B $0.73$ vs.\ $0.58$ ($-0.51$, $p{=}.013$).
The exception is Qwen2.5-0.5B, whose losses are frequency-flat ($+0.01$, $p{=}.97$),
consistent with its lowest baseline competence (0.68) and unreliable inclusion metric. 
Without that conditioning, a Zipf$\times$quantization term over all 429 words
(word-clustered) points the same way in five of these seven artifacts and is significant
in two (Qwen2.5-1.5B and LFM2-1.2B), since it also counts words the model never knew,
which cannot be lost.

\subsection{Vocabulary size does not predict fragility; scale does}\label{sec:vocab}

The fragmentation hypothesis predicts that models with relatively smaller 
vocabulary (32k vs. 131k), which split rare words into more subword
tokens, should degrade more.
It follows from how subword tokenizers are built~\citep{sennrich2016,kudo2018}:
algorithms such as Byte-Pair Encoding start from individual characters and
repeatedly merge the most frequent adjacent symbol pair in the training
corpus into a new token, until the vocabulary reaches its target size.
Since rare words are made of character sequences that co-occur
infrequently, the merges needed to assemble them into a single token are
lower priority and often do not survive the fixed vocabulary budget,
leaving rare words split across more, shorter subword tokens.
Common words, built from frequently co-occurring sequences, get merged early and
typically end up as a single token, or at least fewer tokens than a rare
word of comparable length.

\begin{figure}[!t]
\centering
\includegraphics[width=\textwidth]{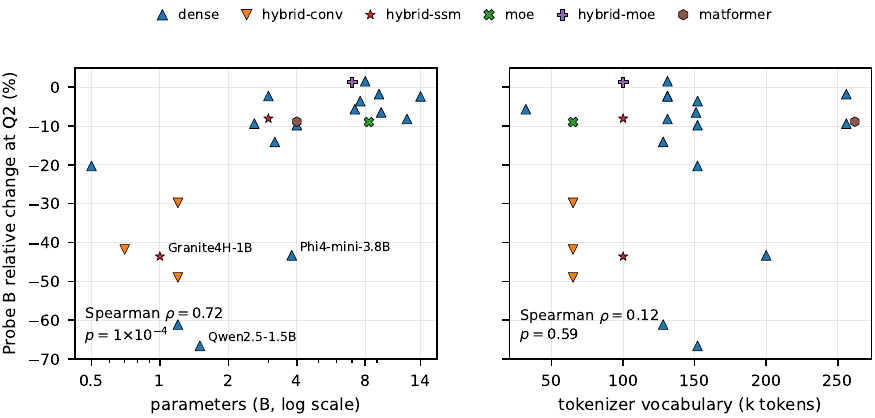}
\caption{Predictors of definitional fragility across the 23 non-collapse
primary artifacts; colour = architecture class. Left: \probeB{} relative
change at Q2 against parameter count (log scale); Spearman $\rho{=}0.72$,
$p{=}1{\times}10^{-4}$. Right: the same quantity against tokenizer vocabulary
size; $\rho{=}0.12$, $p{=}.59$. Parameter count is associated with damage; vocabulary size is not}
\label{fig:predictors}
\end{figure}

\begin{figure}[!t]
\centering
\includegraphics[width=\textwidth]{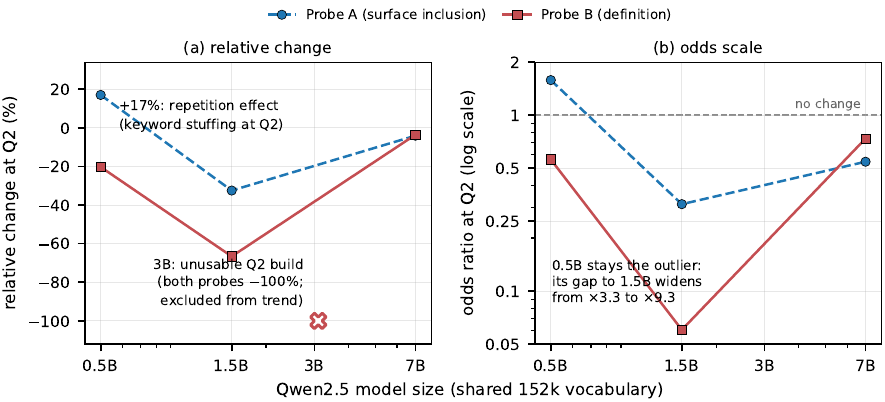}
\caption{Same-tokenizer scale ladder across the Qwen2.5 family (0.5B, 1.5B, 7B;
identical 152k vocabulary), both probes at Q2. (a) Relative change.
(b) The same ladder on the odds scale, which is insensitive to baseline
differences: the 0.5B departure is not an effect of that artifact's low
baseline, since its gap to 1.5B widens from ${\times}3.3$ to ${\times}9.3$.
The 3B point (open cross, panel a) is the collapsed Q2 build, excluded from
the trend; the positive 0.5B inclusion value is the repetition effect of
Table~\ref{tab:main}}
\label{fig:qwenladder}
\end{figure}

\emph{Our data do not support the fragmentation hypothesis, in three ways
(Figs.~\ref{fig:predictors} and~\ref{fig:qwenladder}).}

1. Two dense Mistral-lineage artifacts whose vocabularies differ fourfold,
Mistral-7B (32k) and Mistral-Nemo-12B (131k), are both robust ($-5.7\%$ vs.\
$-8.2\%$ definitional), though they also differ in parameter count and in
baseline precision, the 12B being measured from a Q8 file rather than FP16.

2. Across the 32k vocabulary cohort, Mistral-7B is robust while
Phi-3/3.5 collapse entirely, signalling failure at the artifact level rather than
vocabulary level; and

3. Across all 23 non-collapse primary artifacts with a Q2 build, the Spearman correlation between
vocabulary size and definitional drop is $\rho{=}0.12$ ($p{=}.59$). Conversely,
the correlation for parameter count is very strong:
$\rho{=}0.72$ ($p{=}1{\times}10^{-4}$).

The scale correlation is not an effect of pooling two build providers: 
it holds independently within each build-provider subset of the 23 non-collapse artifacts
(Ollama-native $\rho{=}0.75$, $n{=}10$, $p{=}.013$;
Unsloth $\rho{=}0.66$, $n{=}13$, $p{=}.015$).
The same-tokenizer Qwen2.5 ladder (0.5B/1.5B/7B, all sharing the same 152k tokenizer)
shows severe definitional damage at small scale that
vanishes by 7B ($-66.6\%$ at 1.5B vs.\
$-3.6\%$ at 7B).
(Note: We do not interpret the ladder as strictly monotonic. The 0.5B model's losses are frequency-flat and its inclusion metric is unreliable, so that artifact is not showing the rare-selective damage the
ladder tracks. The 3B artifact is the collapsed build discussed earlier.)

So we could claim that \emph{it is the parameter count, not tokenizer vocabulary,
that predicts definitional robustness under quantization.}

This is not a surprising pattern for us.
A systematic on-device evaluation of GGUF-quantized models had already reported the same
scale-dependence for reasoning and code generation~\citep{ondevice2025}: from Q5\_K to Q3\_K, a 14B model holds its
GSM8K accuracy at $89\%$ while a 1.5B model falls from $59.1\%$ to
$46.5\%$ over the identical precision range, with code generation
degrading even more sharply.
While their results are measured on mathematical reasoning or code correctness,
our result extends the scale-dependence
to lexical-semantic knowledge.
\citet{zhou2026scaling} found a similar threshold when fitting a
scaling law to 293 post-training quantization configurations across
the Qwen3 and Llama-3 families.
They found a phase transition on entry to 2-bit.
Below roughly 2B parameters, performance collapsed across every task type and
the scaling fits stopped converging, while models of 4B and above maintained their
memorization and application scores, provided group size and calibration data were chosen well.
Their group-size term is the nearest analogue to the
K\_S/K\_M distinction we observe at Q4.

The same relationship holds in all three studies: 
larger models are disproportionately more resilient to quantization damage. 
The fact that this relationship reappears across such different tasks, metrics and
quantization severities makes it unlikely to be specific only to reasoning
benchmarks, or to our own probes.

This scale effect is independently corroborated by \citet{dutta2024accuracy}. 
They find that  across different model families (Llama-2, Yi)
and different tasks (MMLU, PIQA, HellaSwag, ARC),
larger models flip fewer individual answers under quantization than smaller ones,
while noting that the residual rate stays non-negligible: 
Llama-2-70B still flips $3$--$8\%$ of answers at 4-bit. 
They attribute it, tentatively, to
larger models being more resistant to the perturbation introduced by quantization.
None of the four studies varied scale while holding the other factors fixed,
so the convergence is consistent with a general effect, but does not establish one.

\subsection{Within-family scale ladders}\label{sec:famscale}

The correlation in the previous subsection is computed \emph{across} model families,
so parameter count as a variable is confounded there with difference in the training data,
instruction-tuning recipe, build provider, and their tokenizers. 
The presented results might raise the question of whether the results we show are because
the model families themselves are weaker. Our experimental sweep is large enough to
answer that directly. 
We have six model families which provide multiple parameter sizes on a fixed tokenizer (for that family), 
covering 18 of the 27 artifacts and 129 file-level
evaluations: 
Granite-4.0-H (350M/1B/3B/7B, 100k vocabulary size), 
Qwen2.5 (0.5B/1.5B/3B/7B, 152k), 
LFM2 (700M/1.2B/8B, 65k), 
Ministral-3 (3B/8B/14B, 131k), 
Llama-3.2 (1B/3B, 128k), and Gemma-2 (2B/9B, 256k). 

Within each of these families, the tokenizer, 
the build provider and the evaluation protocol are held fixed. 

\begin{figure}[!t]
\centering
\includegraphics[width=\textwidth]{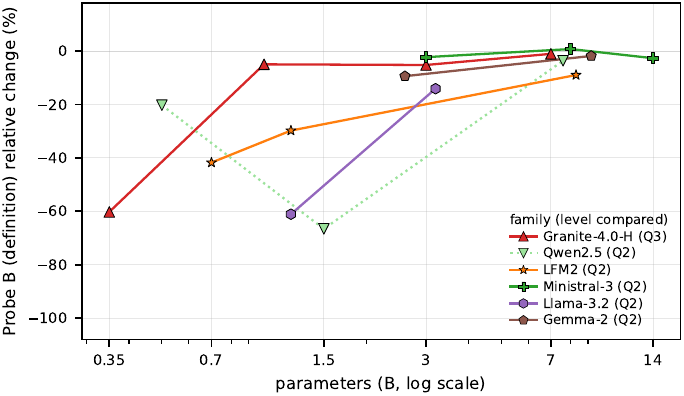}
\caption{Within-family scale ladders: \probeB{} relative change against parameter
count, one line per family (colour and marker shape both distinguish family),
each compared at the most aggressive precision
level all of that family's members were evaluated at (given in the legend;
Granite-4.0-H at Q3 $\ddagger$). Qwen2.5's collapsed 3B build is omitted from
this panel; it appears as the dotted trace in Fig.~\ref{fig:famladders}}
\label{fig:famscale}
\end{figure}

Fig.~\ref{fig:famscale} compares the members of each family at their \emph{most
aggressive} shared precision level used in the study, 
using the worst published sub-variant at that level (out of K\_L, K\_M and K\_S). 
Meaning, across a family of models, such as Ministral-3, we take the most aggressive quantized variant from each of the 
3 models we studied. For example, for Ministral-3 3B/8B/14B we analyse the effect of changing just the parameter size at that quantization level.
Every other result in this paper uses the K\_M file by default.  
We depart from that default here  
to capture the broadest available evidence across all 129 files, 
without restricting six ladders to a single sub-variant.

Within each family we compare its smallest member against its largest at
that matched level (e.g., Gemma2:9B-q2 vs Gemma2:2B-q2).
This gives one comparison per family and six in total,
across the changing \emph{parameter count}.
The \emph{cross-family} correlation predicts that the 
larger member should be less damaged in definitional degradation, which 
we directly check here \emph{within-family}. 

In five of the six families the larger member is the less damaged one.
The single exception is Ministral-3, by less than half a percentage point.
The same five of six hold under each of the three sub-variant conventions:
1. the K\_M file alone, 2. the worst published sub-variant, and 3. the mean over every
published sub-variant.
Changing the sub-variant used for analysis does not affect the overall inference
we are claiming. 

Granite-4.0-H is the clearest instance we show, which has the widest parameter span we studied, 
across a twentyfold range on a single 100k tokenizer.
Its three dense models show a clear size gradient:
the 350M member already loses $60.2\%$ of its
definitional capability by Q3, its 1B member is intact until Q2, and the 3B
member loses $8.1\%$ at Q2.
Its 7B member is not significantly damaged at any level, 
but it changes its architecture to hybrid-MoE.
We cannot claim that last step as parameter count alone, 
as we cannot separate the change in architecture as a confound for it, 
but the pattern holds for the remaining three 350M, 1B and 3B members, 
across a single dense (hyb-SSM) architecture.

\begin{figure}[!t]
\centering
\includegraphics[width=\textwidth]{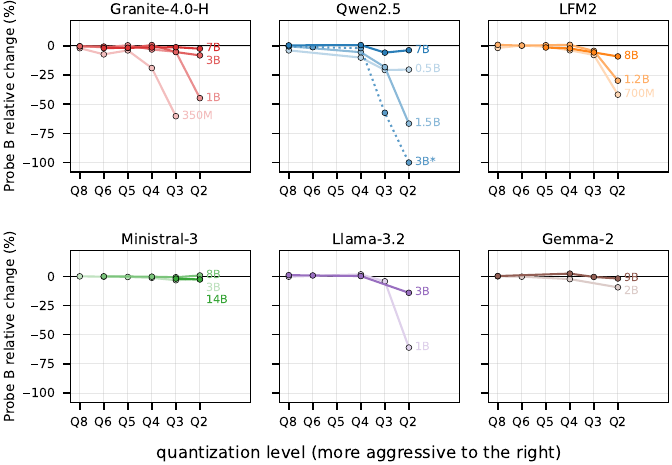}
\caption{The same six families across every evaluated precision level, one panel per
family and one line per member, with darker shading for larger parameter
counts. Each point is the worst sub-variant evaluated at that level. Where the
larger member's trajectory falls below the smaller member's it does so by under
$3.5$ percentage points, except in Qwen2.5 at Q2. The dotted Qwen2.5 trace
(3B$^{*}$) is the collapsed build}
\label{fig:famladders}
\end{figure}

Fig.~\ref{fig:famladders} plots every level for every member
of all the 6 families in this section.
Within a family the larger member's trajectory sits at or
above the smaller member's trajectory at nearly every level,
from which we infer that increasing scale raises the endpoint (across family members at the same level),
as well as provides a later, and shallower descent into semantic degradation.
The two families producing dissimilar curves in Fig.~\ref{fig:famladders} are explained here:

1. Ministral-3's whole curve is almost flat, as none of its members shows meaningful semantic
damage at all, its three artifacts moving by $-2.3\%$, $+0.8\%$ and $-2.7\%$ at Q2. 
And the $-0.4\%$ gap between its 3B and 14B members is negligible.

2. Qwen2.5 does conform to our claim, at its endpoints.
Its 7B member loses $3.6\%$ against the 0.5B
member's $20.3\%$, but it is not monotonic in between, as the 1.5B member loses $66.6\%$.
We have already shown that the 0.5B artifact's losses are frequency-flat and its inclusion metric is not reliable, so it
does not show the rare-selective signature the ladder is following, even though its
definitional loss is real.
Its 3B member is the collapsed build.

We do not claim that these ladders hold everything except parameter count fixed. 
Training data and compute per size are not observable from the released artifacts, 
since even open-source releases publish only weights and architectures.
Two families also change architecture class at their largest
member: Granite-4.0-H at 7B and LFM2 at 8B are both mixture-of-experts
builds activating a fraction of their stored parameters per token
(${\sim}1$B and ${\sim}1.5$B respectively), where every smaller member of both
families is dense.

Two members are also measured from a baseline below full precision, because
no higher-precision build fits within our 16GB VRAM budget: Ministral-3's
14B member from a Q6 file, and Granite-4.0-H's 7B member from a Q8 file.
Using an already-quantized baseline instead of FP16 understates an
artifact's true degradation, since the quality lost between
FP16 and that baseline never enters the measured delta.
For Granite-4.0-H-7B this makes the reported robustness look somewhat
stronger than a full-precision comparison would. 
For Ministral-3 the direction is reversed: it is
already the flatter of the two exception families, 
so the below-FP baseline does not explain its flatness.
Neither caveat is load-bearing on its own.

The narrower conclusion we claim here, which the cross-family correlation could not establish, is that the association between scale and definitional
robustness survives when family, provider and tokenizer are held constant,
in five of the six families we can test it in.

\subsection{Perplexity tracks the damage but cannot calibrate it}\label{sec:ppl}

We tested nine artifact-matched ladders (8 models spanning 0.5--7B in parameter size, including two robust artifacts, one 
unusable build, and two available builds of the same model) to test whether a standard aggregate metric
can calibrate rare-word semantic degradation. 

\begin{figure}[!htbp]
\centering
\includegraphics[width=\textwidth]{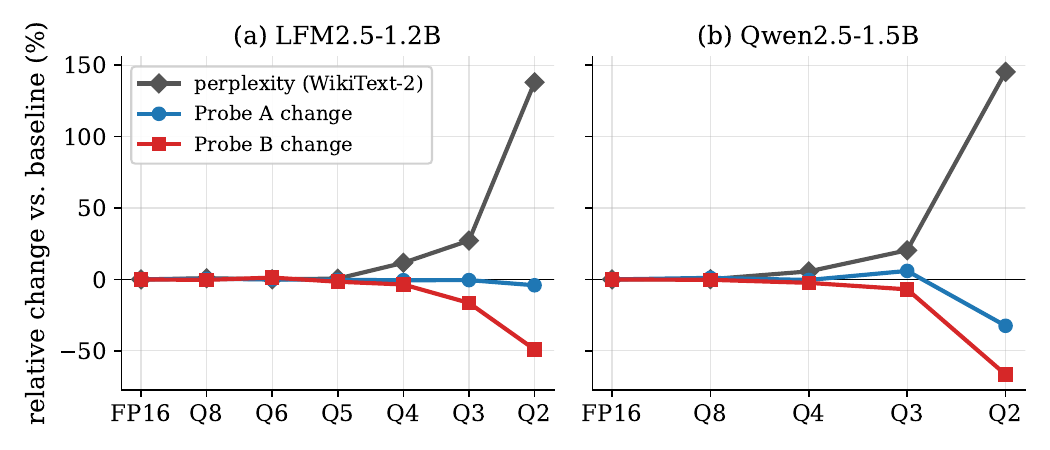}
\caption{Replication of the perplexity comparison of
Fig.~\ref{fig:signatures}c in the two other dissociating sub-2B ladders:
(a) LFM2.5-1.2B, (b) Qwen2.5-1.5B (K\_M files); Perplexity (diamonds) rises
along the familiar accelerating quantization curve while the definitional
probe (squares) drops sharply at Q2; surface inclusion (circles) moves far less}
\label{fig:pplrep}
\end{figure}

\begin{figure}[!htbp]
\centering
\includegraphics[width=\textwidth]{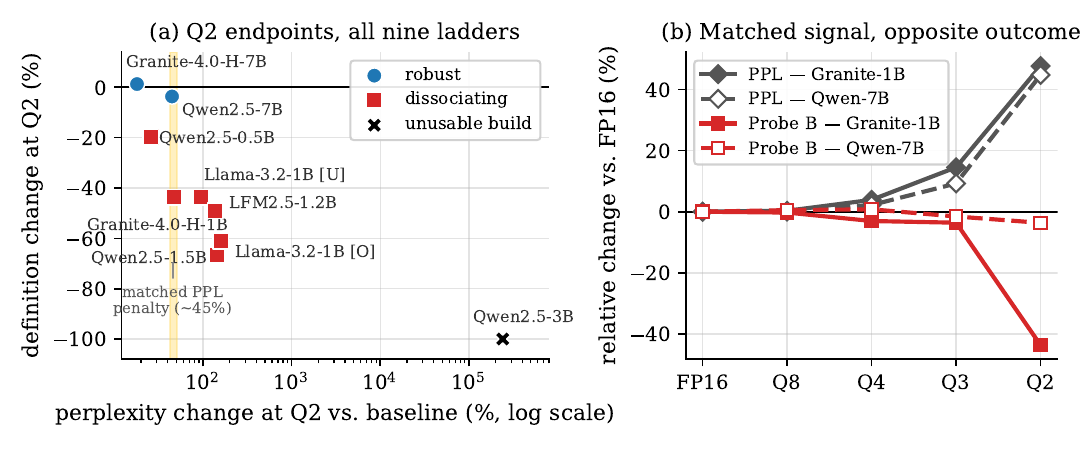}
\caption{The full perplexity program: nine artifact-matched WikiText-2
ladders (K\_M files; Granite-4.0-H-7B from its Q8 baseline). 
(a) Q2 endpoints, perplexity change (log scale) vs.\
relative definitional change. Nearly matched penalties (shaded band) separate the robust Qwen2.5-7B
from the dissociating Granite-4.0-H-1B;
(b) Granite-4.0-H-1B
(filled, solid) vs.\ Qwen2.5-7B (open, dashed) on shared levels:
near-coincident perplexity curves, opposite definitional outcomes}
\label{fig:pplprog}
\end{figure}

WikiText-2~\citep{merity2017wikitext} perplexity
(\texttt{llama-perplexity}, 100 chunks of 512 tokens, identical text at every
level of a ladder, K\_M files) rises along the same familiar accelerating
curve documented in published quantization-quality tables over the 3--8 bit
range they cover~\citep{whichquant2026}, continuing unbroken into Q2 in every
functional ladder, across dissociating as well as robust artifacts
(Figs.~\ref{fig:signatures}c, \ref{fig:pplrep}, and~\ref{fig:pplprog}). 

At Q2 the penalty is $+47.6\%$ for Granite-4.0-H-1B ($11.80{\to}17.42$), $+137.9\%$
for LFM2.5-1.2B, $+145.3\%$ for Qwen2.5-1.5B, $+25.9\%$ for Qwen2.5-0.5B. 
Crucially, the penalty is noticeable for the robust models as well:
$+44.7\%$ for the robust Qwen2.5-7B ($7.71{\to}11.16$) and
$+18.0\%$ for the robust Granite-4.0-H-7B MoE (from its Q8 baseline).

The perplexity curve is not flat, but it cannot be read as a measure of definitional
damage, in three specific ways.

    1. \emph{Matched Penalties, Opposite Outcomes}:
    Qwen2.5-7B and Granite-4.0-H-1B suffer nearly identical Q2 perplexity penalties ($+44.7\%$
    vs.\ $+47.6\%$) while losing $3.6\%$ vs.\ $43.6\%$ of their definitions, respectively:
    a twelvefold difference in lexical damage from the same perplexity signal
    (Fig.~\ref{fig:pplprog}b). 
    On the odds scale each artifact is measured against its own \probeB{}
    baseline ($0.897$ vs.\ $0.844$), not against \probeA{} as in the
    probe$\times$quantization interaction odds ratio.
    Qwen2.5-7B's odds fall from $8.71$ to $6.41$, a ratio of $0.74$, against
    Granite-4.0-H-1B's $5.41$ to $0.91$, a ratio of $0.17$: a $4.4$-fold gap. 
    The direction and the qualitative result are the same on both scales. 
 
    2. \emph{Matched Damage, Unmatched Penalties}: 
    In contrast, Granite-4.0-H-1B and the Unsloth build of Llama-3.2-1B lose the same
    $43.6\%$ of definitions (both define 362 of 429 words at FP16 and 204 at Q2, though
    not the same words) at $+47.6\%$ vs.\ $+93.7\%$ perplexity,
    a twofold divergence in penalty for the same lexical outcome.
    The $\Delta$PPL${\to}\Delta$definition trade-off at Q3 spans wildly, from
    $+9.2\%{\to}{-}1.6\%$ (Qwen2.5-7B) to $+70.1\%{\to}{-}47.1\%$ (Qwen2.5-3B).
    
    3. \emph{No Usable Absolute Threshold}: 
    Absolute perplexity doesn't provide any baseline for \emph{semantic retention}.
    LFM2.5-1.2B at Q3 (PPL 24.8) defines $60.4\%$
    of the rare words while Qwen2.5-1.5B at Q2 (PPL 25.0) defines just $29.1\%$. 
    Within a single tokenizer, where absolute perplexities are directly
    comparable, Qwen2.5-3B at Q3 (PPL 15.38) defines $47.1\%$ while Qwen2.5-0.5B
    at FP16 (PPL 15.29) defines $67.8\%$: the almost same aggregate metric describing a
    quantization-damaged artifact as well as a merely small healthy one. 

Across the nine Q2 endpoints, perplexity change and definitional loss
show strong rank-correlation (Spearman $\rho{=}0.96$; $0.95$ excluding the
collapsed build): both worsen together as overall damage grows more severe.
This correlation however does not provide a useful threshold to detect the definitional damage.
Neither the absolute perplexity nor its change tells a practitioner how much
definitional knowledge a given build has lost.
However, perplexity does flag the unusable quantized builds clearly.
The collapsed Qwen2.5-3B Q2 build scores
PPL 21{,}797 against a 9.04 baseline. 
Beyond flagging unusable builds, perplexity cannot tell a practitioner that the ability to define rare words is what was lost.

Perplexity, as an aggregate likelihood over ordinary text, is dominated by
common tokens; it tracks generic degradation, 
but is not designed to isolate
the selective loss of a narrow, low-frequency capability.
This limitation is partly inherent to the metric itself: perplexity is computed as
the inverse of the geometric mean of per-token probabilities, 
so a handful of catastrophically low-probability tokens making up 
a lost rare-word definition in our study 
can be mathematically cancelled by higher probabilities elsewhere in the same
pass~\citep{dutta2024accuracy}.

A practitioner selecting a quantization level based on standard perplexity tables
will see Q2 as generically \emph{worse}, but they cannot infer why it is worse, specifically if
they want to use the model for tasks which require rare-word knowledge and capabilities.
Perplexity cannot reveal the exact threshold where a model forgets the meaning of rare words,
nor can it signal which model has relatively better semantic
capabilities compared to another model with the same score.

\subsection{Artifact-source sensitivity}\label{sec:source}

\begin{figure}[!t]
\centering
\includegraphics[width=0.6\textwidth]{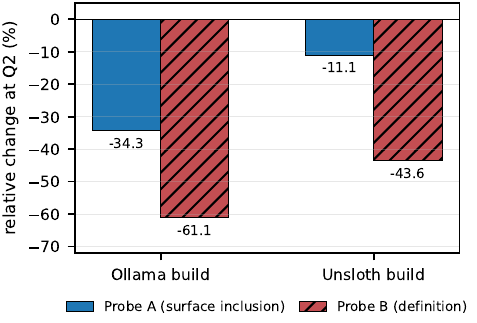}
\caption{Artifact-source sensitivity: relative change at Q2 on both probes
for Llama-3.2-1B in two independently quantized GGUF builds of the same
model (Ollama vs.\ Unsloth). The definitional loss's direction replicates, while
magnitude differs across builds}
\label{fig:source}
\end{figure}

Quantized builds of the same model from different providers are not
equivalent (Fig.~\ref{fig:source}). 
Llama-3.2-1B in Ollama's build loses
$34.3\%$/$61.1\%$ (inclusion/definition) at Q2; the independent Unsloth build
loses $11.1\%$/$43.6\%$. 
On the odds scale, which we are using since the
two builds' baselines differ ($0.821$ vs.\ $0.844$), the same asymmetry
holds: Ollama's odds fall from $4.59$ to $0.47$, a ratio of $0.10$, against
Unsloth's $5.41$ to $0.91$, a ratio of $0.17$ (identical to Granite-4.0-H-1B's by coincidence).
The build sensitivity is therefore not merely a result of using a percentage-change scale.
Nor does it begin at Q2. At FP16 only 63 of the 429 definitions are identical across the
two builds, so part of what separates them is present before quantization, and our design
cannot separate it from the quantization pipeline. The builds also diverge at Q4\_K\_M,
where Unsloth's loses $5.0\%$ ($p{=}.010$) and Ollama's gains $1.7\%$.

The larger loss in the definitions than inclusion is 
replicated across both builds, but the magnitude of the loss is different. 

Our regime classification of a file under study is based on the
artifact, not models. 
The strata of Table~\ref{tab:main} classify artifacts as well. 
This difference between build providers affects how a model is classified. 
For example, Llama-3.2-1B's Ollama build falls in Regime IIb, while its Unsloth
build falls in Regime~II proper, its inclusion falling
$11.1\%$ against $43.6\%$ on definitions.

We found that the dissociation is not
specific to a single provider's quantization pipeline.
Our sweep itself spans both providers: the sub-2B cohort contains Ollama-native
artifacts (Llama-3.2-1B, Qwen2.5-0.5B/1.5B) and Unsloth artifacts
(Granite-4.0-H-1B, the LFM2 family).

Build sensitivity also replicates on the aggregate metric for Llama-3.2-1B: the two builds'
FP16 files score identical WikiText-2 perplexity (15.14), yet at Q2 Ollama's
build costs $+158.0\%$ ($15.14{\to}39.07$) against Unsloth's $+93.7\%$
($15.14{\to}29.34$), which is directionally aligned with their definitional losses
($-61.1\%$ vs.\ $-43.6\%$). 
So perplexity does pick up the difference between the two builds.
Although this evaluation rests on a single model we used for calibration across two
build providers, 
we can only infer about directionality, not magnitude of definitional loss.
Our data do not support the use of one provider's perplexity
loss to quantitatively predict or proxy for definitional loss of an alternate provider's
build of the same model. 

The behaviour of a quantized model therefore must be validated at the artifact level,
not the architecture level.

\section{Discussion}\label{sec:discussion}

\subsection{Why saying survives while knowing fails}

GGUF K-quants~\citep{llamacpp} use different effective precisions for different tensors:
they preserve token-embedding and output
matrices at higher effective bit-widths while 
compressing interior FFN, attention, or state-space blocks. 
Producing a word supplied in the prompt (\emph{surface inclusion}) depends mostly on
embedding and output-matrix precision, which GGUF's scheme protects, plus the local context. 
A definition, however, is not retrieved directly as a stored string;
it must be
constructed at inference time by composing semantic associations \emph{encoded as
distributed representations} across the model's interior weights. If that information is more
vulnerable to quantization, definitional performance could
fall while surface inclusion remains relatively intact.

The frequency results fit this interpretation. 
Low-frequency words receive relatively less exposure during pre-training,
so their semantic representation plausibly contains less redundancy and
would therefore be more vulnerable to weight perturbations. 
Quantization noise could then disrupt their semantic reconstruction first, while the
common-word control, whose associations are reinforced by far more training exposure,
survives the same noise.

An independent, similar result points the same way from a different task. 
\citet{hua2026bias} found that quantization changed answers
selectively based on how confident the model was about them.
Items the full-precision model was least confident about flipped three to eleven
times more often after quantization than items it was confident about.
Rare words are also likely to be low-confidence items, so a process that preferentially
damages low-confidence items would reach rare words first, 
without needing any lexically specific mechanism.
Their probe is a bias benchmark and ours is a lexical task, so this is corroborating
evidence about \emph{which} items are fragile. 
Sec.~\ref{sec:limits} considers the same selectivity at the decision level,
where confidence becomes decision margin.

Under the precision-allocation interpretation, our two probes benchmark opposite ends of the lexical pipeline:
\probeA{} measures surface-form production (which is protected), while
\probeB{} measures the integrity of meaning construction (which is not protected).
This structural reason might explain why the
dissociation runs in the direction it does and not the reverse. 
The dissociation appears across architectures with very different interior blocks:
attention, Mamba-2 SSM, hybrid-convolution. GGUF uses the same weight-block quantization scheme for
all three, while state-space models have needed separate, 
architecture-specific quantization methods~\citep{quamba2024}.
A generic effect of low-precision weight perturbation on the distributed semantic
associations is therefore a more plausible explanation than a failure tied to one
particular architecture.

We did not inspect the affected weights directly,
however, so this remains an interpretation of the behavioural results rather
than a mechanistic finding.

\subsection{Reconciliation with feature-level compression studies}

\citet{duan2026} finds that sparse-autoencoder features of a
full-precision model survive RTN quantization in a graded rather than cliff-like fashion.
Most features that do not survive RTN quantization are degraded (\emph{blurred}) rather than damaged (\emph{destroyed}); 
which features survive can be predicted from their full-precision activation statistics;
and perplexity can miss the feature damage entirely. 
Superficially, this differs from our finding that quantization can
substantially reduce rare-word definition performance.
The possible explanation
we propose, as a hypothesis, is compositional: 
individual features can remain usable even when their
combination becomes less reliable. 
A definition, however, requires several pieces of
semantic information to be combined during generation.
Retrieval of a single feature survives blurring; reconstruction of a
precise meaning from many simultaneously blurred, thinly-reinforced features fails first. 
This interpretation would also explain why the definitional failure concentrates on rare
words, which have the fewest redundant supports, and why surface inclusion is spared, since it
leans on protected tensors and local context rather than deep composition.
Perplexity can also
miss the effect, since common tokens contribute much more to its aggregate
value.
Testing whether quantization
actually disrupts feature combinations would require analysis of the model's
internal representations, which we leave for future work
(Sec.~\ref{sec:future}).

At the representational level, \citet{wang2026compressed} 
traced quantization-induced loss of factual recall to the last layers of the network,
using attribution at both the neuron and layer level. 
They found that the relations most affected were those where
full-precision performance had not saturated.
We interpret low-saturation knowledge as thinly encoded, the same property we
attribute to rare words, argued from the representational side rather than the
behavioural one.
Their measured losses are modest and they conclude that quantization is still worth doing.
Our results are compatible with theirs. 
They evaluated 4-bit and 8-bit builds from GPTQ, AWQ and integer quantization,
and state that 1-bit and 2-bit quantization is outside their scope. 
Our sharpest dissociations occur at Q2 and Q3, below that range.
Even at the bit-widths they tested,
five of our artifacts show significant definition loss at Q4, all at or below 4B,
smaller than any model they tested.
This is consistent with their finding that smaller models within a family lose more.

\subsection{Stored capacity may matter more than active compute}

A parameter-count only interpretation of the scale effect is complicated by 
our two Mixture-of-Experts (MoE) artifacts. 
MoE architectures by design contain a large number of total parameters,
but a single token activates only a small subset of those during
auto-regressive inference. 

The two MoE artifacts in our sweep are both robust through Q2, despite active
compute comparable to the dense sub-2B models that show severe dissociation.
LFM2-8B-A1B (8.3B stored, ${\sim}1.5$B active per token) loses only
$\dB{=}{-}9.0\%$ (interaction $p{=}.46$), and Granite-4.0-H-7B (7.0B stored,
${\sim}1$B active) gains $\dB{=}{+}1.3\%$ (interaction $p{=}.006$) from a Q8
baseline, which understates its degradation relative to a full-precision
comparison (Sec.~\ref{sec:famscale}).
In both models the interaction coefficient is positive on the log-odds scale
(Table~\ref{tab:main}): definitional accuracy holds up at least as well as
surface inclusion once \probeA{}'s near-ceiling baseline is accounted for,
though only Granite-4.0-H-7B's coefficient is individually significant.

This resilience suggests that stored capacity protects 
semantic knowledge independent of active compute, the two quantities
which normally get confounded in dense
architectures.
It aligns with the redundancy hypothesis above as well, that
a rare-word association must survive the destruction of
its redundant copies, whose number scales with total stored parameters rather than
with the parameters active in a single forward pass.

\citet{allenzhu2025capacity}
provide a quantitative baseline for our qualitative hypothesis:
transformer language models are estimated to have 
a knowledge capacity of 2 bits per parameter;
quantization to int8 shows negligible impact on
this capacity, while int4 reduces it to roughly 0.7 bits per parameter.

Two further results from their study support our behavioural observation:

    1. Mixture-of-experts models retain nearly all of their theoretical capacity
    despite activating only a small fraction of their parameters per token
    (activation sparsity).
    With 32 experts, capacity falls only 1.3-fold relative to the base dense
    scaling law in their 1{,}000-exposure setting, and 1.5-fold at 100
    exposures, even though just 8.8\% of total parameters are active during
    inference.
    If capacity were dependent only on active parameters, that 8.8\%
    figure would imply a much larger reduction, roughly eleven-fold.
    The observed reduction therefore supports the premise that knowledge
    capacity tracks stored rather than active parameters. They do not quantize
    any mixture-of-experts model, so whether that stored capacity is what
    protects our own MoE artifacts under quantization remains our inference,
    not their result.
    
    2. Separately, they find that knowledge seen only 100 times during
    training, what they term ``\emph{100-exposure}'', is stored at half the
    bit-per-parameter density of knowledge encountered 1{,}000 times:
    1 bit per parameter rather than the 2 bits their sufficiently-trained
    models reach.
    Less-exposed knowledge occupies less capacity independent of quantization, 
    the same low-exposure, low-redundancy logic proposed above for rare words.

The perplexity ladder of Granite-4.0-H-7B completes this picture. 
Its Q2 perplexity rises only $18.0\%$ ($9.09{\to}10.73$ from its Q8 baseline), the
smallest penalty among our nine ladders, and its definition score changes
minimally by $+1.3\%$ as well.
This is despite having active compute of just ${\sim}1$B, the scale
at which the 1--1.5B artifacts in the same ladder set all dissociate and pay
perplexity penalties of $+47.6\%$ to $+158.0\%$
(Sec.~\ref{sec:ppl}, Fig.~\ref{fig:pplprog}a).
On both metrics, definitional degradation and aggregate perplexity,
this MoE artifact's behaviour is more consistent with its larger
stored capacity (total parameters), rather than being limited by
the vulnerabilities of small active compute.

This should not be read as a general claim that larger models are more robust.
\citet{hua2026bias} evaluated Qwen2.5 from 0.5B to 14B and found no consistent
scaling advantage for bias stability: the 14B artifacts were no more stable than
the 0.5B ones, and the within-family pattern was erratic.
Their measurements stop at 4-bit and concern alignment behaviour rather than
stored knowledge, so the two findings do not conflict.
Taken together, these findings show that scale protection is capability-specific:
our evidence is strictly constrained to the preservation of lexical-semantic knowledge at the K-quant floor.

\subsection{Practical guidance}

For practitioners deploying GGUF artifacts on consumer hardware, we recommend the
following four guidelines: 

1. 
Q4\_K\_M is lexically stable for every artifact of 1B or larger in our sweep
except one, Unsloth's build of Llama-3.2-1B, which loses $5\%$.
The more aggressive Q4\_K\_S variant already starts to show small losses in 
some artifacts, and below 1B two of the three artifacts already
lose material definitional accuracy at Q4.

2. 
At precisions below Q4, small ($\lesssim$2B) models should be assumed to lose
definitional knowledge even when output remains fluent. 
Applications depending on lexical precision, such as technical writing, 
education, translation, or domain-specific assistants,
should validate the models at the artifact level. 
This is particularly relevant for on-device or edge deployments, 
which are more likely to run smaller models on consumer hardware.

3.
Unusable Q2 builds can ship without any warning.
To screen for this before
deployment, we recommend using either a 100-word common-word probe or a single
WikiText-2 perplexity check.
Both are cheap and catch unusable build failure. 
Unusable builds fail even on common words (as confirmed directly in
one case, Phi-3.5), and perplexity rises far beyond any functional build in
the one Regime-I artifact we measured a ladder for (${\sim}2{,}400{\times}$
baseline), while every Regime-II artifact in our nine-ladder sweep
remained within functional range.

4. 
Perplexity catches unusable builds, but it is not a substitute for
capability-specific validation.
In our nine perplexity ladders, we observed that nearly identical Q2 perplexity
penalties ($+44.7\%$/$+47.6\%$) correspond to both intact ($-3.6\%$) and severely
dissociated ($-43.6\%$) definitional outcomes. 
No perplexity threshold, on either its absolute value or its change, 
identifies the selective definitional damage.
This blindness is not specific to perplexity or to our probes; 
for instance, aggregate bias scores conceal response-level flips by averaging
them out~\citep{hua2026bias}.
Hua et al. recommend a stratified, task-specific post-quantization evaluation in
place of benchmark averages, and our results support the same for lexical semantics.

\newpage

\section{Limitations}\label{sec:limits}

\textbf{Scorer and metric.} 
Our scorer is a string matcher, not an LLM judge;
we bound the matcher's
error with an LLM-judge census of every definition (primary judge gpt-oss-20b, replicated
by Gemma~4; its misses are quantization-asymmetric) and release the word lists, synonym
key, evaluation code, every definition with both judges' verdicts, both
human-verification samples (for the synonym key and the judges), and
all ${\sim}100$k word-level scored definition outcomes at
\url{https://github.com/NietzscheDostoevsky/snk-release}.
The word list (Gemini 3.1 Pro, Google), the synonym key (Claude Fable~5, Anthropic) and the scorer's
validation (gpt-oss-20b, OpenAI; Gemma~4, Google) all involve LLMs, which we disclose as a
potential source of circularity. The primary judge's developer has no other role in the
pipeline, and each judge agrees with a human author on 62 of 65 items;
the human verification of 65 definitions and of 20 synonym-key entries is released with the
reviewer's per-item verdicts, and both checks can be re-read from released data.
The rare-word set's rarity claim rests on corpus statistics (Zipf frequency) alone.
The binary inclusion metric is
insensitive to overuse (Qwen2.5-0.5B inflates its Q2 score by
repetition, which we flag).
\probeA{} is deliberately the lighter task, and at full precision it scores
above \probeB{} in all but two of the 27 artifacts (both Qwen2.5;
Fig.~\ref{fig:gallery}). Our conclusions rest on the probe contrast and the
controls, and do not require the two probes to be equally hard.

\textbf{Decision margins and selectivity.} 
The probe asymmetry can also be viewed in terms of \emph{decision margin}.
\citet{dutta2024accuracy} show that quantization
disproportionately flips low-margin decisions and that harder tasks flip
more than easier ones; \citet{hua2026bias} find the same
item-level selectivity from a different confidence statistic.
Both our probes supply the target word; what differs is what we are asking the
model to generate in response to those probes.
\probeA{} scores the surface inclusion of a token \emph{already} in the context, which can
be argued as a relatively high-margin operation, whereas \probeB{} requires the
model to construct and produce that word's stored meaning unaided,
a comparatively lower-margin task.
Margin offers a plausible candidate mechanism for the inclusion/definitional dissociation, i.e.,
``rare-word meaning is thinly encoded'' and
``rare-word definitions are low-margin decisions'' describe the same phenomenon at two levels of description, 
and margin might be how thin encoding shows up at the decision boundary. 
Our behavioural finding that fluent production of words the artifact
can no longer define holds under either reading, and the common-word
control already rules out the pure task-difficulty version of the account.
Our result establishes the behavioural dissociation, but this design cannot 
tell us whether the underlying explanation should be described at the 
representation level (``thin encoding'') or decision level (``low margin'').
The distinguishing test is a per-token log-probability analysis, with margins measured directly on
both probes across the ladder, which our pipeline did not capture and which
we leave to future work (Sec.~\ref{sec:future}).

\textbf{Evaluation ecosystem and corpora.} 
Scope varies across controls and comparisons: 
the common-word control covers 11 of the 27 artifacts (including all eight sub-2B models), 
while perplexity comparisons span nine ladders across eight models (0.5B--7B), 
encompassing two robust artifacts and one unusable build.
Each probe uses a single prompt template, 
and perplexity evaluations rely on a single corpus (WikiText-2) and 
a single measurement protocol (\texttt{llama-perplexity}, 512-token chunks).
Prompt sensitivity evaluation and multi-corpus generalization, 
alongside perplexity protocol sensitivity control, remain future work. 

All artifacts were served through the Ollama/llama.cpp inference ecosystem,
and originate from two quantization providers (Ollama-native and Unsloth).
One model in the sweep was evaluated in both providers' builds.
Degradation direction was consistent across the two builds, but magnitude differed.
From this single comparison we can infer the effect's directionality but not its magnitude across build providers.

To address mutable registry tags and potential upstream re-quantization, 
exact registry tags for all evaluated artifacts are logged in the released package.
Cryptographic digests were not captured during initial probe sweeps due to 
local storage retention constraints; 
however, the released package includes SHA-256 digests 
for 46 of the 51 perplexity files, covering eight of the nine ladders (excluding Qwen2.5-1.5B).

\textbf{Linguistic scope.} 
The study is English-only. 
Several sweep families (Qwen2.5, Mistral-Nemo, Gemma-3n) 
are explicitly multilingual, so part of their trained capacity also
serves languages that our probes do not test. 
Also, whether the rare-word dissociation observed in English extends
to other languages, particularly lower-resource languages, remains untested.

\textbf{Scale-severity anomalies.} 
The relationship between model scale and quantization severity is
not strictly monotonic within a family (Sec.~\ref{sec:famscale}).
On the shared Qwen2.5 tokenizer the 1.5B member loses more of its definitions
at Q2 than the smaller 0.5B member does. 
The same lineage's 3.1B member is unusable at Q2 and is excluded
from the scale trend for that reason (Sec.~\ref{sec:collapse}). 
As with the other Regime-I artifacts, we cannot distinguish a quantizer error in that
build from a capacity threshold specific to that model at this compression
ratio. The probe outputs alone do not favour either explanation.

\section{Future Work}\label{sec:future}

The behavioural evidence localizes the damage to meaning construction rather
than surface inclusion. 
The natural next step is to identify which internal structures fail, 
and the present results specify what these experiments should test, 
beginning with three directions of increasing evidential strength.

First, 
\emph{embedding-geometry tracking}: 
measure the geometry of rare-word
representations across the quantization ladder in the models already studied,
requiring instrumentation rather than new sweeps. It would establish whether
the artifacts that lose definitions are the ones whose rare-word geometry
moves, and whether the words lost are those that move furthest.

Second, 
a \emph{predictive signature}: 
link a representational statistic
computed on the FP16 model to which specific words lose their definitions
after quantization, in the spirit of the survival predictors of \citet{duan2026}. 
It could provide a word-level prediction of which words
would survive quantization, and up to what level of aggressive compression.

Third, 
the \emph{causal step}: 
protect candidate fragile directions during quantization and show definitions recover, or
inject selective perturbations and show definitions break while surface
production is untouched. 
Qwen2.5-3B at Q3 could be used as a natural testbed: a dense,
HF-loadable artifact with a large, clean behavioural dissociation on fluent output
($0.92{\to}0.83$ inclusion vs.\ $0.89{\to}0.47$ definition) and a
same-tokenizer family ladder for scale contrasts. 
Before such internal measurements, however, 
the dequantized model should first be verified to reproduce the behavioural 
dissociation observed in the present evaluation.

A fourth
direction addresses the decision-margin question:
re-running both probes through an inference pipeline that logs per-token
output probabilities would let us compute decision margins and
quantization-induced flips in the spirit of \citet{dutta2024accuracy}.
This would test whether the definitions lost
under quantization are disproportionately low-margin decisions, and could provide a
mechanistic account of the dissociation that the current log-probability-free
Ollama runs cannot establish. 
Existing margin and uncertainty results~\citep{dutta2024accuracy,hua2026bias} are
computed by ranking a supplied set of answer options. 
\probeB{} has no option set, so the test proposed here is per-token margins over free
generation, scored against a reference definition rather than against a chosen option.

A fifth direction would extend \probeA{} from \emph{production} to
\emph{usage}.
Verbatim inclusion establishes that the word form is produced, but does not establish 
whether it is deployed with its intended sense.
Separating the two requires a
usage-quality assessment on the generated sentences. 
Since appropriate use and grammatical correctness are subject to expert judgement 
rather than deterministic scoring, 
a rigorous usage-quality evaluation requires qualified annotators,
an agreement statistic across at least two of them, and a word-paired sample
large enough to support the comparison.
A narrower, deterministic proxy is available for one
part of it: malformed morphological variants of the target word
(\emph{ephemerous} for \emph{ephemeral}, \emph{idiosyncracy} for
\emph{idiosyncrasy}) can be detected by dictionary lookup against WordNet and
the \texttt{web2} word list, which were used as independent sources to validate the
stimulus set used here as well. This would
provide a reproducible measure of word-form corruption without requiring
judgement about semantic appropriateness.

Finally, 
the evaluation itself can be broadened naturally to multilingual lexicons, 
additional GGUF sources, prompt-template robustness, 
and perplexity generalization across additional corpora and measurement protocols.

\section{Conclusion}\label{sec:conclusion}
Across 27 quantized artifacts spanning four architecture backbones,
aggressive GGUF quantization does not degrade lexical competence uniformly.
Below ${\sim}2$B parameters, it can split definitional knowledge from fluent
surface production, leaving models that continue to produce rare words in their output, usually at
near-baseline rates, after losing the ability to define them.
The damage is rare-word-selective and
frequency-graded, begins earlier in the quantization ladder
as the model size decreases, is not
calibrated by perplexity, 
and varies between different provider builds of the same model.
Total collapse, where it occurs, is a property of individual quantized builds
rather than an intrinsic property of rare-word semantics.
Tokenizer vocabulary size does not predict it (Spearman $\rho{=}0.12$).
Model capacity and training exposure, not the token inventory, 
are the probable constraints on the retention of meaning. 
The validation burden for an AI practitioner 
shifts from architectures to artifacts, and the central lesson is
\emph{fluency is not evidence of meaning}.

\section*{Data and code availability}

The word lists (with Zipf validation), the multi-synonym answer key, all
${\sim}100$k word-level scored definition outcomes with per-configuration
inclusion metrics, per-level statistics (McNemar,
interaction, onset grid, within-family scale ladders), every definition with both
judges' verdicts, the human-verification samples, and the perplexity
ladders (nine ladders over eight models, including a dual-build pair) are
available at
\url{https://github.com/NietzscheDostoevsky/snk-release}.

The tiered scorer, the judge rubric and runner, a script that reproduces the
scorer-validation numbers of Sec.~\ref{sec:scoring}, and the word-list validation
script are included in the same repository. The inference notebooks that
produced the generations are not released; the prompts, decoding parameters
and procedures they implement are specified in the Methods section.

\bibliography{references}
\bibliographystyle{tmlr}

\end{document}